\documentclass[runningheads]{llncs}

\usepackage{eccv}

\usepackage{eccvabbrv}

\usepackage{graphicx}
\usepackage{booktabs}

\usepackage[accsupp]{axessibility}  

\usepackage{booktabs}
\usepackage{amsmath}
\usepackage{epsfig}
\usepackage{multirow}
\usepackage{ dsfont }
\usepackage{ cite }
\usepackage{float}
\usepackage{color, colortbl, xcolor}
\usepackage{soul}
\usepackage{float}
\usepackage{url}
\usepackage{afterpage}
\usepackage{rotate}
\usepackage{makecell}
\usepackage{multirow}
\usepackage{algorithm}
\usepackage{algorithmicx}
\usepackage{algpseudocode}
\usepackage{mathtools}
\usepackage{etoolbox}
\usepackage{lipsum}
\usepackage{bbm}
\usepackage{listings} 
\usepackage{tabularx}
\usepackage{amssymb}
\usepackage{amsbsy}
\usepackage{pifont} 
\usepackage{tikz}
\usetikzlibrary{shapes.geometric}
\usepackage[normalem]{ulem}
\usepackage{array}
\usepackage{transparent}
\usepackage[dvipsnames]{xcolor}
\usepackage{bbm}

\usepackage{wrapfig}
\definecolor{2nd}{gray}{0.8}

\newcolumntype{g}{>{\columncolor{Gray}}c}

\newcolumntype{^}{>{\currentrowstyle}}

\algdef{SE}[SUBALG]{Indent}{EndIndent}{}{\algorithmicend\ }%
\algtext*{Indent}
\algtext*{EndIndent}

\DeclareMathOperator*{\argmax}{arg\,max}

\newcommand{\redtick}{\textcolor{red}{\ding{52}}}

\newcolumntype{Y}{>{\centering\arraybackslash}X}
\newcolumntype{P}[1]{>{\centering\arraybackslash}p{#1}}

\algnewcommand{\nComment}[1]{\Statex \Comment{#1}}
\definecolor{LightCyan}{rgb}{0.88,1,1}

\definecolor{mypurple}{RGB}{153, 0, 153}
\definecolor{mygray}{RGB}{128, 128, 128}
\definecolor{mygreen}{RGB}{0, 153, 0}

\definecolor{mycyan}{RGB}{64, 128, 128}
\definecolor{mypink}{RGB}{255, 182, 193}
\definecolor{myred}{RGB}{165,42,42}
\definecolor{myyellow}{RGB}{255, 191, 0}

\definecolor{tab_red}{rgb}{0.71, 0.11, 0.0}
\definecolor{tab_green}{rgb}{0.11, 0.71, 0.0}

\newcommand{\improve}[1]{\textcolor{tab_green}{\textbf{+#1}}}
\newcommand{\drop}[1]{\textcolor{tab_red}{\textbf{-#1}}}

\newcommand{\thickhline}{%
	\noalign {\ifnum 0=`}\fi \hrule height 1pt
	\futurelet \reserved@a \@xhline
}
\makeatletter
\global\let\oriCT@@do@color\CT@@do@color

\usepackage[breaklinks,colorlinks]{hyperref}

\usepackage{orcidlink}

\begin{document}

\title{CPM: Class-conditional Prompting Machine for Audio-visual Segmentation}

\titlerunning{Class-conditional Prompting Machine (CPM)}

\author{
Yuanhong Chen\inst{1}\orcidlink{0000-0002-8983-2895} \and
Chong Wang\inst{1} \and
Yuyuan Liu\inst{1} \and
Hu Wang\inst{2} \and
Gustavo Carneiro\inst{3}
}

\authorrunning{Y.~Chen et al.}

\institute{
Australian Institute for Machine Learning, University of Adelaide, Australia \and
Mohamed bin Zayed University of Artificial Intelligence, United Arab Emirates \and
Centre for Vision, Speech and Signal Processing, University of Surrey
\email{yuanhong.chen@adelaide.edu.au}
}

\maketitle

\begin{abstract}
Audio-visual segmentation (AVS) is an emerging task that aims to accurately segment sounding objects based on audio-visual cues. The success of AVS learning systems depends on the effectiveness of cross-modal interaction. Such a requirement can be naturally fulfilled by leveraging transformer-based segmentation architecture due to its inherent ability to capture long-range dependencies and flexibility in handling different modalities. However, the inherent training issues of transformer-based methods, such as the low efficacy of cross-attention and unstable bipartite matching, can be amplified in AVS, particularly when the learned audio query does not provide a clear semantic clue. In this paper, we address these two issues with the new \underline{C}lass-conditional \underline{P}rompting \underline{M}achine (CPM). CPM improves the bipartite matching with a learning strategy combining class-agnostic queries with class-conditional queries. The efficacy of cross-modal attention is upgraded with new learning objectives for the audio, visual and joint modalities. We conduct experiments on AVS benchmarks, demonstrating that our method achieves state-of-the-art (SOTA) segmentation accuracy\footnote{This project is supported by the Australian Research Council (ARC) through grant FT190100525.}.

\end{abstract}

\section{Introduction}
\label{sec:intro}

The recognition and integration of auditory and visual data are fundamental to human cognitive processes, playing a critical role in facilitating meaningful communication~\cite{murray2011neural}. Audio-visual segmentation (AVS) is an emerging cross-modal reasoning task that mimics such cognitive processes, aiming to localize visual objects based on audio-visual cues.
AVS has many applications, such as 
the automatic localisation and identification of sounding objects to improve the accessibility of videos for the blind and visually impaired~\cite{liu2021makes}.
A major challenge in AVS is achieving effective cross-modal interaction between sound and visual objects~\cite{senocak2023sound}. 
Many AVS methods~\cite{zhou2022audio,mao2023multimodal,chen2023closer} generally adopt a traditional per-pixel classification framework~\cite{wu2022language,cheng2022masked}, utilising early fusion strategies (e.g., cross-attention fusion ~\cite{vaswani2017attention}) together with an FCN decoder~\cite{long2015fully,chen2018encoder} to make predictions. 
Such per-pixel design has achieved good performance, but it tends to under-utilise the audio data due to its lower informativeness compared with the visual data~\cite{senocak2023sound, chen2023closer}. 
Contrastive learning can mitigate this issue by matching informative audio-visual pairs~\cite{chen2023closer}. However, another critical limitation of the per-pixel design is its failure to capture instance-level visual information, resulting in inconsistent segmentation predictions within or between frames in a video sequence~\cite{wu2022language}.


\begin{figure}[t]
    \centering    \includegraphics[width=1.0\linewidth]{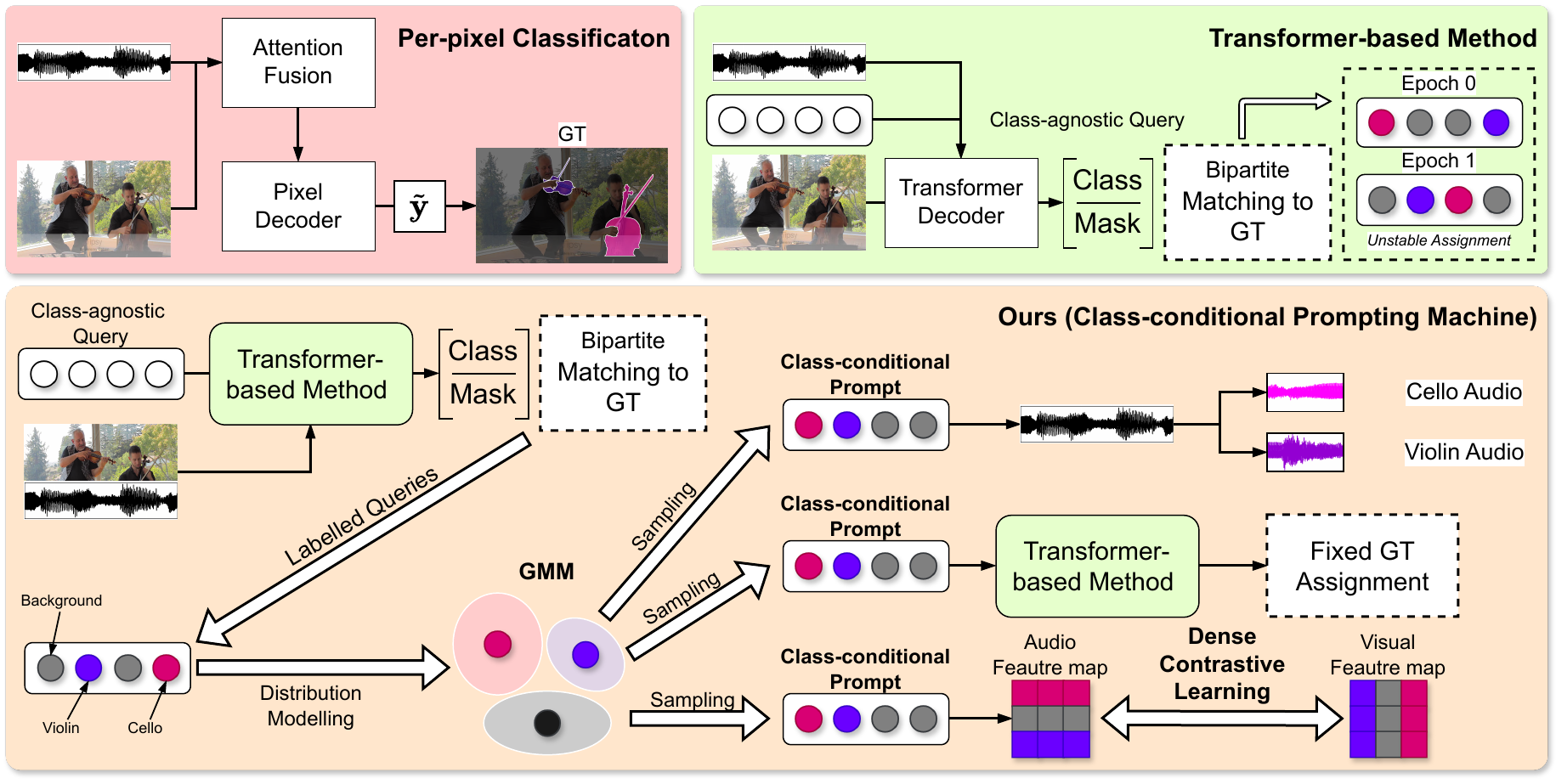}
    \vspace{-10pt}
    \caption{Comparing conventional AVS methods~\cite{chen2023closer,gao2023avsegformer} with our CPM approach, CPM inherits the class-agnostic query from transformer-based methods and integrates class-conditional prompts sampled from the learned joint-modal data distribution to achieve three objectives: 1) learn disentangled audio partitioning, 2) facilitate semantic-guided object identification, and 3) promote more explicit audio-visual contrastive learning.
    }
    \vspace{-10pt}
    \label{fig:motivation}
\end{figure}

These two issues have been addressed by transformer-based AVS methods designed to capture instance-level information and to rely on more effective contrastive learning~\cite{wu2022language,liu2023gres,liu2023audio,li2023catr,liu2023bavs,gao2023avsegformer,yang2023cooperation}.
Nevertheless, these AVS approaches still show slow convergence and relatively poor accuracy~\cite{liu2022dab,li2022dn,zhang2023mp}, primarily attributed to the \textit{low efficacy of the cross-attention}~\cite{sun2021rethinking} and the \textit{unstable bipartite matching}~\cite{li2022dn}. 
Solutions for these problems are based on integrated masked-attention~\cite{cheng2022masked}, which focuses on features around predicted segments, and anchor de-noising~\cite{li2022dn} to reconstruct the noisy ground-truth bounding box detection.
However, these solutions may not work well in AVS because of the weak constraint provided by global audio features that contain a mixture of sound sources~\cite{gao2023avsegformer,liu2023bavs,li2023towards} resulting in more instability during training. 
It is worth noting that all methods above have the common issue of relying on class-agnostic prompts that provide little guidance to the bipartite matching process, thereby reducing training efficacy. 

In this paper, we introduce the \underline{C}lass-conditional \underline{P}rompting \underline{M}achine (CPM), an audio-visual segmentation training approach that leverages class-conditional prompts to enhance bipartite matching stability and improve cross-modal attention efficacy.
To enhance bipartite matching, we introduce a novel learning method, combining class-agnostic queries~\cite{li2023towards,cheng2022masked,gao2023avsegformer} with class-conditional queries, sampled from our iteratively updated generative model of class-specific embeddings, with the former queries matched to ground-truth labels using the Hungarian Algorithm~\cite{carion2020end}, followed by individual processing of class-conditional queries in two modalities.
To improve cross-modal attention efficacy, new learning objectives are proposed for both audio and visual modalities. In the audio modality (audio conditional prompting, denoted as ACP), the original spectrogram is corrupted with auditory off-the-screen noise and class-conditional queries are employed to reconstruct the original spectrogram, while in the visual modality (visual conditional prompting, denoted as VCP), noisy class-conditional queries sampled from our generative model are used to probe semantically similar content in the image space.
To further upgrade the performance of cross-modal attention, %
we introduce a new prompting-based audio-visual contrastive learning (PCL) task, guided by class-specific queries to densely constrain the cross-modal representations. 
To summarise, our main contributions are:
\begin{itemize}
    \item A new AVS training approach to enhance bipartite matching stability and enhance the efficacy of cross-modal attention. Our core innovation lies in the development of a \underline{\textbf{C}}lass-conditional \underline{\textbf{P}}rompting \underline{\textbf{M}}achine (CPM).
    \item To improve the bipartite matching, we propose a new AVS learning strategy that combines class-agnostic queries with class-conditional queries, sampled from our iteratively updated generative model of class-specific embeddings.
    \item The efficacy of cross-modal attention is upgraded with new learning objectives for the audio, visual and joint modalities.
    For \textbf{audio}, we present ACP that perturbs the original spectrogram and uses class-conditional queries to reconstruct the spectrogram; for \textbf{visual}, we introduce VCP that explores noisy class-conditional queries sampled from our generative model to probe the corresponding semantic in visual feature map. To enhance the cross-modal attention efficacy further, we propose PCL, consisting of a new prompting-based \textbf{joint} (audio-visual) contrastive learning task.
\end{itemize}

We firstly show the effectiveness of our CPM model through rigorous evaluation on established benchmarks such as AVSBench-Objects~\cite{zhou2022audio} and AVSBench-Semantics~\cite{zhou2023audio}. Furthermore, we extend our evaluation by including VPO synthetic benchmarks~\cite{chen2023closer}, aiming to enhance our comprehension of AVS methods' capacity to capture audio-visual correlations. Our findings across these benchmarks consistently demonstrate that our approach yields better classification accuracy compared to existing methods.

\section{Related Works}
\label{sec:related_work}
\subsection{Transformer-based architecture}

Transformer-based methods have shown promising performance in detection~\cite{carion2020end} and segmentation~\cite{cheng2021per, cheng2022masked} benchmarks. The fundamental concept is to leverage the object query to probe image features from the output of transformer encoders and bipartite graph matching to perform set-based box/mask prediction. It is also evidenced that such a framework can benefit multi-modal learning due to its attention mechanism and flexibility in data modelling~\cite{xu2023multimodal}. 
Despite its successful application in various domains~\cite{carion2020end,cheng2022masked}, such frameworks have also been reported to exhibit a poorer convergence rate compared to traditional CNN-based methods~\cite{li2022dn, liu2022dab, cheng2022masked}. In exploring the reasons behind the poor convergence, previous studies have emphasized enhancing the interpretability of the learnable query~\cite{meng2021conditional,liu2022dab} (often referring to them as anchor points/boxes). 
Alternatively, the utilisation of denoising methods~\cite{li2022dn,zhang2023mp} has also been adopted to facilitate the learning of bounding box offset~\cite{li2022dn} and improve the utilisation of the transformer decoder layers~\cite{zhang2023mp}. Furthermore, masked attention has proven effective in enhancing both convergence rate and model performance~\cite{cheng2022masked}. 
The successful implementation of denoising methods and masked attention mechanisms has inspired us to devise a mitigation strategy for bipartite matching that promotes improved cross-modal understanding within the cross-attention process.

\vspace{-10pt}
\subsection{Audio-visual Segmentation (AVS)} 
Audio-visual Segmentation is a dense classification task or detection of sounding visual objects in videos, using image sequences and audio cues. Zhou et al.~\cite{zhou2022audio} introduced the AVSBench-Object and AVSBench-Semantics benchmarks~\cite{zhou2023audio}, which enable the evaluations of single-source and multi-source AVS tasks for salient and multi-class object segmentation. 
To address concerns related to the high annotation cost and dataset diversity, Chen et al.~\cite{chen2023closer} introduced a cost-effective strategy to build a relatively unbiased AVS dataset, named Visual Post-production (VPO).
Mainstream AVS methods use audio as the reference query~\cite{liu2023audio,huang2023discovering, gao2023avsegformer,li2023catr,liu2023annotation,liu2023bavs,liu2023audiovisual,chen2023closer} or prompt~\cite{mo2023av,wang2023prompting}. For example, some methods adopt MaskFormer~\cite{cheng2021per} or segment anything model (SAM)~\cite{kirillov2023segment} to perform image segmentation using audio queries or encoded audio prompts and cross-attention layers. These methods benefit from the attention mechanism's ability and mask-classification's features to capture long-range dependencies and enhance image segmentation ability~\cite{cheng2021per}, spatial-temporal reasoning~\cite{li2023catr} and task-related features~\cite{li2023catr,liu2023audiovisual,liu2023bavs,gao2023avsegformer,zhou2022audio, mao2023multimodal}.
The adaptation of the Maskformer-based framework for AVS relies on methods to encourage the audio-visual semantic alignment~\cite{liu2023bavs,li2023catr,yang2023cooperation,liu2023audio}.
Such strategy mitigates the poor audio semantic information caused by modality imbalance~\cite{chen2024bootstrapping} and learns disentangled multi-modal representations~\cite{li2023towards,yang2023cooperation} via vector quantization~\cite{gray1984vector} and bidirectional attention mechanism. 
Even though the query-based transformer architecture~\cite{carion2020end} has shown great success in semantic segmentation tasks, certain weaknesses such as low efficacy of cross-attention and unstable bipartite matching process~\cite{li2022dn} are still only partially addressed. These two issues may be exacerbated in  AVS due to the poor audio semantic information.

\vspace{-10pt}
\subsection{Audio-visual Contrastive Learning}
Contrastive Learning has shown promising results in audio-visual learning (AVL) methods~\cite{chen2021localizing,hu2022mix,mo2022localizing,mo2022closer,chen2023closer}. These methods bring together augmented representations from the same instance as positives while separating representation pairs from different instances as negatives within a batch. A reported issue with current AVL contrastive learning is its reliance on self-supervision~\cite{chen2020simple} to connect audio and visual representations of the same class. To overcome this issue, CAVP~\cite{chen2023closer} propose a supervised contrastive learning~\cite{khosla2020supervised, wang2021exploring, li2022targeted} method that mines informative contrastive pairs from arbitrary audio-visual pairings to constrain the learning of audio-visual embeddings. 
Nevertheless, these previous approaches predominantly leverage global audio representations, thereby limiting the model's capacity to discern individual audio sources. 
Consequently, this constraint reduces the model's effectiveness in scenarios involving multiple sound sources.
In our work, we employ class-specific queries to retrieve the corresponding audio representations, which facilitates a more explicit form of contrastive learning between audio and visual modalities.

\section{Method}
\label{sec:method}
We denote a multi-class audio-visual dataset as $\mathcal{D}=\{(\mathbf{a}_i, \mathbf{x}_i, \mathbf{y}_i, \mathbf{t}_i))\}_{i=1}^{|\mathcal{D}|}$, where $\mathbf{x}_i\in\mathcal{X}\subset \mathbb{R}^{H \times W \times 3}$ is an RGB image with resolution $H \times W$, $\mathbf{a}\in\mathcal{A}\subset \mathbb{R}^{T \times F}$ denotes the magnitude spectrogram of the audio data with $T$ time and $F$ frequency bins, $\mathbf{y}_i\in\mathcal{Y}\subset \{0, 1\}^{H \times W \times C}$ denotes the pixel-level ground truth for the $C$ classes (the background class is included in these $C$ classes), and $\mathbf{t}_i\in\mathcal{Y}\subset \{0, 1\}^{C}$ is a multi-label ground truth audio annotation. We use Mask2former~\cite{cheng2022masked} as the segmentation framework.

\vspace{-10pt}
\subsection{Preliminaries using Class-agnostic Queries}
\label{sec:method_prelim}
\begin{figure}[t]
    \centering    \includegraphics[width=1.0\linewidth]{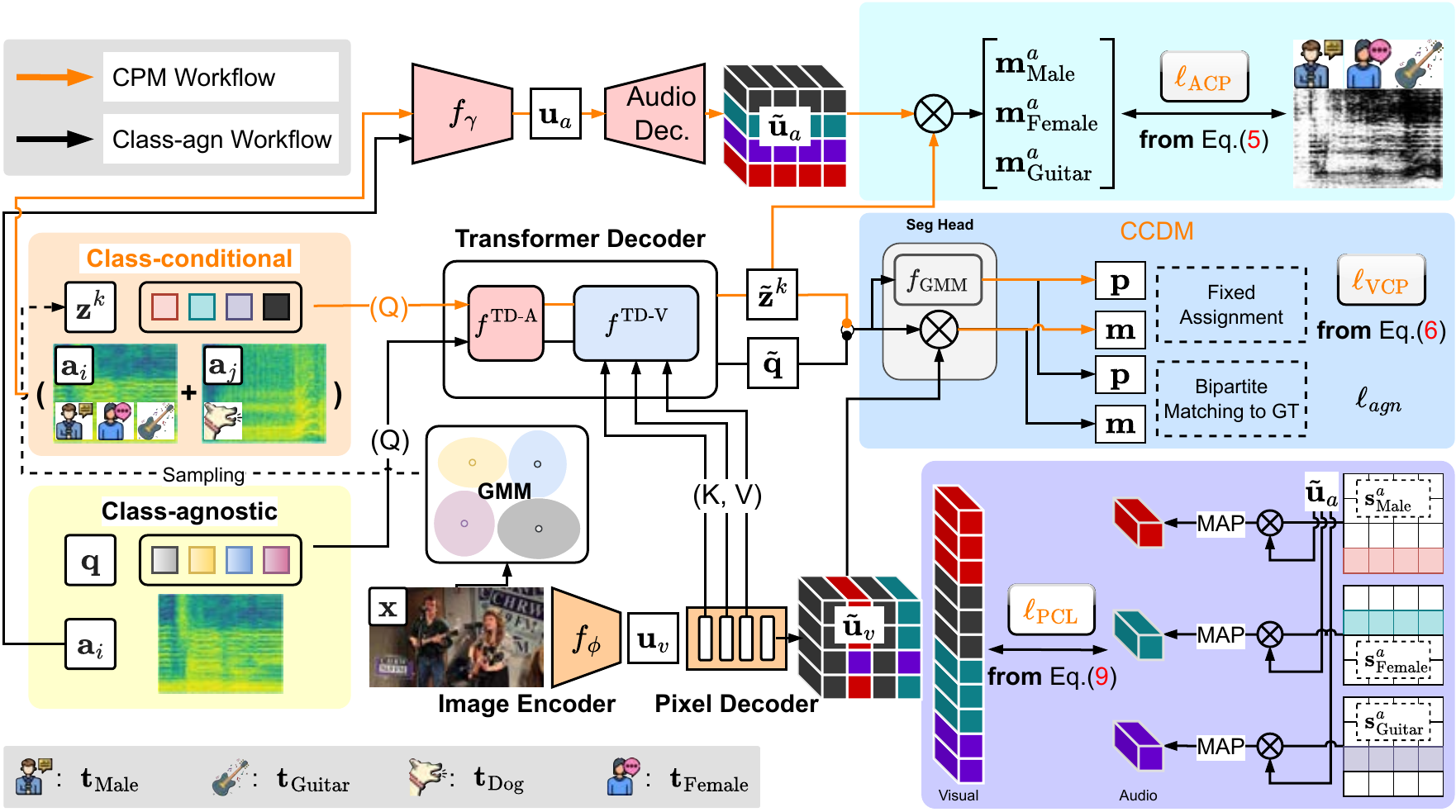}
    \caption{Illustration of our CPM method. Starting with a scene with a mixture of sound sources including \textcolor{red}{Male}, \textcolor{teal}{Female} and \textcolor{purple}{Guitar}, 
    the training alternates between the use of learnable class-agnostic queries, and queries sampled from
    a list of class-specific query features from the GMM, denoted as $\mathbf{z}^{\text{male}}$, $\mathbf{z}^{\text{female}}$ and $\mathbf{z}^{\text{guitar}}$. 
    The overall training objective is composed of three learning tasks: 1) in 
    audio conditional prompting (ACP), we aim to use $\mathbf{z}^{\text{male}}$, $\mathbf{z}^{\text{female}}$ and $\mathbf{z}^{\text{guitar}}$ to recover the original magnitude spectrogram $\mathbf{a}_i$ from the noise spectrogram (i.e., $\mathbf{a}_i$ + $\mathbf{a}_j$) that is corrupted by another \textcolor{brown}{Dog} audio signal $\mathbf{a}_j$; 2) a visual conditional prompting (VCP) that aim to probe the corresponding pixels w.r.t to the class-specific query features; and 3) a contrastive learning task that target to densely constrain the audio and visual representations. 
    For training, both the CPM Workflow, indicated by the \textbf{\textcolor{orange}{orange arrow}}, and the Class-agnostic Workflow, marked by the \textbf{\textcolor{black}{black arrow}}, are utilized. However, only the Class-agnostic Workflow is used for inference.
    }
    \vspace{-10pt}
    \label{fig:frame_cpm}
\end{figure}

Like other Maskformer-based methods~\cite{wu2022language,liu2023bavs,gao2023avsegformer,liu2023audio}, we aim to learn the parameters $\theta \in \Theta$ for the model $f_{\theta}:\mathcal{X} \times \mathcal{A} \to [0,1]^{H \times W \times C}$, which comprises the image and audio backbones that extract features with $\mathbf{u}_a=f_\gamma(\mathbf{a})$ and $\mathbf{u}_v=f_\phi(\mathbf{x})$, respectively, where $\gamma,\phi \in \theta$, and $\mathbf{u}_a,\mathbf{u}_v \in \mathcal{U}$, with $\mathcal{U}$ denoting a unified feature space. 
A set of learnable query features (comprising the object query feature and positional query embeddings) are defined as the joint-modal output embeddings similar to the Perceiver model~\cite{jaegle2021perceiver}. We define the class-agnostic query feature as $\mathbf{q}\in\mathcal{Q}\subset\mathbb{R}^{N \times D_q}$, where $N$ denotes the number of class-agnostic queries, and $D_q$ represents the dimensionality of the feature space. 
As depicted in Fig.~\ref{fig:frame_cpm}, given $\mathbf{q}$, we aim to group pixels with matched semantic information from $\mathbf{u}_a$ and $\mathbf{u}_v$ through consecutive transformer decoder layers~\cite{vaswani2017attention,cheng2022masked} and generate $N$ mask embeddings $\tilde{\mathbf{q}}\in\mathcal{Q}$ and pixel embeddings $\tilde{\mathbf{u}}_v\in\mathbb{R}^{H\times W\times D_q}$.
Then, the model independently predicts the embeddings into $N$ set of class predictions (via $\mathsf{Softmax}(\cdot)$) and mask predictions (via $\mathsf{Sigmoid}(\cdot)$) denoted as $\mathcal{E}_{\mathsf{pred}}=\{(\mathbf{m}_i,\mathbf{p}_i)\}_{i=1}^{N}$,
where $\mathbf{m} \in \mathcal{M} \subset\{0, 1\}^{H \times W}$ and $\mathbf{p} \in \mathcal{P}\subset\{0, 1\}^{C}$. 
We denote the ground-truth set derived from the training set $\mathcal{D}$ with $\mathcal{E}_{\mathsf{gt}}=\{(\mathbf{m}^{\mathsf{gt}}_i,\mathbf{p}^{\mathsf{gt}}_i)\}_{i=1}^{N^{\mathsf{gt}}}$. 
During training, we use the Hungarian algorithm to perform optimal matching between $\mathcal{E}_{\mathsf{pred}}$ and $\mathcal{E}_{\mathsf{gt}}$~\cite{carion2020end,cheng2022masked} to facilitate the label assignment. Since $|\mathcal{E}_{\mathsf{pred}}| > |\mathcal{E}_{\mathsf{gt}}|$, we pad $\mathcal{E}_{\mathsf{gt}}$ with no-object class $\varnothing$ to achieve one-to-one matching~\cite{cheng2022masked}.

The loss $\ell$ to train the model $f_{\theta}(.)$ includes a cross-entropy query classification loss $\ell_{ce}$, and a binary mask loss $\ell_{mask} = \ell_{focal}+\ell_{dice}$, which combines a focal loss $\ell_{focal}$ and a dice loss $\ell_{dice}$~\cite{cheng2022masked,liu2023bavs,gao2023avsegformer}. The overall training loss for the class-agnostic query feature is defined as $\ell_{agn} = \ell_{ce} + \ell_{mask}$.
During testing, $\mathbf{p}$ and $\mathbf{m}$ are merged via the multiplication $\argmax_{i:c_i\neq\varnothing}\mathbf{p}(c_i)\cdot\mathbf{m}_i$, where $c_i$ is the class label with maximum likelihood $c_i=\argmax_{c\in\{1,...,C,\varnothing\}}\mathbf{p}_i(c)$ for each probability-mask pair indexed by $i$.

\vspace{-10pt}
\subsection{Class-conditional Prompting Machine (CPM)}
The proposed CPM training builds upon the class-agnostic training method from Sec.~\ref{sec:method_prelim} 
by introducing a new training stage based on class-conditional prompting. This allows the probing of both the magnitude spectrogram and the image feature map, aiming to mitigate unstable training issues and improve cross-attention efficacy.
In some cases, the prompts can be manually crafted (i.e., text label, bounding box) based on domain knowledge or specific task requirements~\cite{kirillov2023segment}.
Prompts can also be automatically learned to form a fixed set of prototypical embeddings for each class~\cite{mo2023audio}. 
However, it is impossible to manually define class conditional prompts in high-dimensional latent spaces, and using a limited-size set of learned prompts may fail to capture the comprehensive distribution of class-specific prompts.  
Hence, it is desirable that these class-specific prompts can be sampled from a generative model~\cite{liang2022gmmseg,chen2023generative}.
that comprehensively represents the respective class.
We adopt the Gaussian Mixture Models (GMMs) as such generative model~\cite{reynolds2009gaussian}, which improves the intra-class variability and increases robustness to class imbalances when compared to the alternative approaches mentioned before.
Before delving into the methodology of the CPM, we first introduce the generation process of the class-conditional query features.

\vspace{-10pt}
\subsubsection{Class Conditional Distribution Modelling (CCDM)}
Instead of constructing a posterior $p(c|\tilde{\mathbf{q}})$ with $\mathsf{Softmax}$ classifier for the mask embeddings $\tilde{\mathbf{q}}$, the generative classifier employs the Bayes rule for label prediction, estimating the class-conditional distribution $p(\tilde{\mathbf{q}}|c)$ alongside the class prior $p(c)$~\cite{liang2022gmmseg}, with:
\begin{equation}
    p(c|\tilde{\mathbf{q}}) = \frac{p(\tilde{\mathbf{q}}|c)p(c)}{\sum_{c'} p(\tilde{\mathbf{q}}|c') p(c')},
\label{eq:gmm_posterior}
\end{equation}
where the class probability prior $p(c)$ is uniform. 
To enable GMM modelling in Maskformer architecture~\cite{cheng2021per,cheng2022masked}, we replace the $\mathbf{p}=\mathsf{Softmax}(\tilde{\mathbf{q}})$ activation function with $f_{\mathsf{GMM}}(\tilde{\mathbf{q}})=p(c|\tilde{\mathbf{q}})$ defined in~\eqref{eq:gmm_posterior}
which maps the input mask embeddings $\tilde{\mathbf{q}}$ to probability data density %
over $C$ classes.
Assuming that after the Hungarian matching between the predictions set
$\{(\mathbf{m}_i,\mathbf{p}_i)\}_{i=1}^{N}$ and $\mathbf{y}$, we get a new one-to-one matched ground-truth set $\tilde{\mathbf{y}}$ w.r.t each prediction by padding the $\mathbf{y}$ with no-object class $\varnothing$.
Subsequently, we can form the GMM training dataset $\mathcal{F}=\{(\tilde{\mathbf{q}}_{n},\tilde{\mathbf{y}}_{n})\}_{n=1}^{F}$ by pairing the mask embeddings with the assigned label.
In our method, the goal of GMM is to model the data distribution of the joint-modal mask embedding $\tilde{\mathbf{q}}$ for each class $C$ in the $D_q$-dimensional space by employing a weighted mixture of $M$ multivariate Gaussians, defined as follows:
\begin{equation}
    p(\tilde{\mathbf{q}}|c) = \sum_{m=1}^{M}p(m|c;\boldsymbol{\pi}_c)p(\tilde{\mathbf{q}}|c,m;\boldsymbol{\mu}_c,\mathbf{\Sigma}_c) = \sum_{m=1}^{M}\boldsymbol{\pi}_{cm}\mathcal{N}(\tilde{\mathbf{q}};\boldsymbol{\mu}_{cm},\mathbf{\Sigma}_{cm}),
    \label{eq:gmm_classifier}
\end{equation}
where $\sum_{m}\pi_{cm} = 1$ represents
the mixing coefficients, $\boldsymbol{\mu}_{cm}\in\mathbb{R}^{D_q}$ denotes the mean vector, and $\mathbf{\Sigma}_{cm}\in\mathbb{R}^{D_q\times D_q}$ is the covariance matrix. 
The optimisation of the GMM parameters is performed by the Expectation Maximisation (EM) algorithm~\cite{dempster1977maximum}. In the \textbf{E-step}, we iterate through $\mathcal{F}$, computing the responsibilities using the current parameters of the GMM for each class $c$:
\begin{equation}
    \gamma_{c, n}^{(t)}(m) = \frac{\boldsymbol{\pi}_{cm}^{(t-1)} \mathcal{N}(\tilde{\mathbf{q}}_n; \boldsymbol{\mu}_{cm}^{(t-1)}, \mathbf{\Sigma}_{cm}^{(t-1)})}{\sum_{m' = 1}^{M} \boldsymbol{\pi}_{cm'}^{(t-1)} \mathcal{N}(\tilde{\mathbf{q}}_n; \boldsymbol{\mu}_{cm'}^{(t-1)}, \mathbf{\Sigma}_{cm'}^{(t-1)})}.
\label{eq:e-step}
\end{equation}
We re-estimate the parameters using the calculated responsibility in the \textbf{M-step}.
\begin{equation}
\begin{split}
    \boldsymbol{\mu}_{cm}^{(t)} &= \frac{1}{N_{cm}^{(t)}}\sum_{(\tilde{\mathbf{q}}_{n},\tilde{\mathbf{y}}_{n})\in\mathcal{F}_c}\gamma_{c, n}^{(t)}(m)\;\tilde{\mathbf{q}},
    \quad
    \boldsymbol{\pi}_{cm}^{(t)} = \frac{N_{cm}^{(t)}}{|\mathcal{F}|},
    \\
    \boldsymbol{\Sigma}_{cm}^{(t)} &= \frac{1}{N_{cm}^{(t)}}\sum_{(\tilde{\mathbf{q}}_{n},\tilde{\mathbf{y}}_{n})\in\mathcal{F}_c}\gamma_{c, n}^{(t)}(m)\;(\tilde{\mathbf{q}}_{n}-\boldsymbol{\mu}_{cm}^{(t)})(\tilde{\mathbf{q}}_{n}-\boldsymbol{\mu}_{cm}^{(t)})^\mathsf{T},
\end{split}
\end{equation}
where $N_{cm}^{(t)}=\sum_{(\tilde{\mathbf{q}}_{n},\tilde{\mathbf{y}}_{n})\in\mathcal{F}_c}\gamma_{c, n}^{(t)}(m)$. We maintain an external memory bank containing GMM training samples, defined above with the set $\mathcal{F}$, and the GMM parameters are robustly estimated with momentum~\cite{liang2022gmmseg}.

\vspace{-10pt}
\subsubsection{Audio Conditional Prompting (ACP)} \label{sec:aret_process}
Motivated by the concept of ``mix-and-separate''~\cite{gao2019co,hu2022mix}, we design the ACP process to improve cross-attention interaction for the dense audio feature representations.
We consider 
an off-the-screen noise
dataset $\mathcal{D}_{r} = \{(\mathbf{a}_{j})\}_{j=1}^{|D_{r}|}$. 
For each training iteration, we sample a clean audio sample $\mathbf{a}_i$ from the $\mathcal{D}$ and 
an off-the-screen noisy audio sample $\mathbf{a}_{j}$ from $\mathcal{D}_{r}$ to build the mixture of magnitude spectrogram $\mathbf{a}_{p}=\mathbf{a}_{i}+\mathbf{a}_{j}$. 
The whole ACP process is illustrated in the first section of Fig.~\ref{fig:frame_cpm}.
Before delving into the audio recovery process, we first sample a set of class-conditional prompts 
$\mathbf{z}^{k}\sim \{ \mathcal{N}(\boldsymbol{\mu}_{k},\boldsymbol{\Sigma}_{k})\}_{k \in \mathcal{K}}$ (where $\mathcal{K}=\{k|\mathbf{t}_{i}(k)=1\}$ represent the indices for ground truth labels)
via the GMM model from the last iteration according to the target semantic classes $\mathbf{t}_{i}$ that we want to recover.
The major process of the audio recovery process shares a similar concept with the visual part described in Sec.~\ref{sec:method_prelim}; the difference is that we use the joint-modal mask embeddings $\tilde{\mathbf{z}}^{k}$ to search the semantically similar sound source on the decoded dense audio feature map $\tilde{\mathbf{u}}_a$ and generate their corresponding mask predictions 
$\{\mathbf{m}_{k}^{a} | k\in\mathcal{K}\}$.

The learning objective of the ACP process is to encourage the consistency between the predicted masks $\mathbf{m}_{k}^{a}$ with the derived ground-truth spectrogram ratio mask $\frac{\mathbf{a}_i}{\mathbf{a}_{p}}$.
The AVS datasets generally do not provide the separated sound source data for each semantic class (i.e., in Fig.~\ref{fig:frame_cpm}, we do not have the ground-truth audio data of \textcolor{red}{Male}, \textcolor{teal}{Female} and \textcolor{purple}{Guitar}), hence we adopted a summation operation over the predicted masks to integrate all mixed sound sources. Finally, a Sigmoid activation function $\mathsf{\sigma(\cdot)}$ is applied to constrain the prediction into a binary mask.
We denoted the mean squared error (MSE) loss of the ACP process as follows:
\begin{equation}
    \ell_{\text{ACP}} = \Bigg\|\sigma \left (\sum_{k\in\mathcal{K}}\mathbf{m}_{k}^{a} \right ) - \frac{\mathbf{a}_i}{\mathbf{a}_{p}}\Bigg\|_{2}.
    \label{eq:loss_ACP}
\end{equation}

\vspace{-10pt}
\subsubsection{Visual Conditional Prompting (VCP)}
The VCP module is processed simultaneously with ACP. The design of VCP aims to bypass bipartite matching via the generated class-conditional prompts 
$\mathbf{z}^{k}$
to ease the model training as it can provide a more stable learning target for each query feature to mitigate the instability brought by the bipartite matching with the class-agnostic queries.
Given a set of class-conditional prompts derived from the ground-truth image labels, our training objective of VCP is to correctly segment the corresponding image regions as well as successfully classify these prompts after the consecutive transformer decoder layers. The loss function is similar to $\ell_{agn}$ in Sec.~\ref{sec:method_prelim}, where we replace $\mathbf{q}$ with $\mathbf{z}^{k}$, denoted as $\ell_{\text{VCP}}=\ell_{ce}+\ell_{mask}~\refstepcounter{equation}(\theequation)\label{eq:loss_VCP}$.

\vspace{-10pt}
\subsubsection{Prompting-based Contrastive Learning (PCL)} The ability to learn discriminative feature representation is critical for the audio-visual system. One limitation of the previous audio-visual contrastive learning methods~\cite{mao2023contrastive,chen2023closer} is that they can only leverage the global audio representation due to the lack of class conditional prompts for class-specific feature disentanglement. By taking advantage of class-conditional distribution modelling, we can overcome this limitation by utilising the predicted spectrogram saliency mask $\mathbf{s}_{k}^{a}=\sigma(\mathbf{m}_{k}^{a})$ and its associated class label $k$ of each sound source, denoted as
$\{\mathbf{s}_{k}^{a}|\mathbf{s}_{k}^{a}\in\mathcal{M}\subseteq[0, 1]^{T \times F}, k\in\mathcal{K}\}$. 
To disentangle the class-specific representations for the audio feature map, we iteratively apply the masked average pooling (MAP)~\cite{siam2019amp,zhou2022regional} to $\mathbf{u}_{a}$ 
based on $|\mathcal{K}|$ saliency masks $\mathbf{s}_{k}^{a}$ for the feature extraction process. We first threshold $\mathbf{s}_{k}^{a}$  and convert it to a binary mask via $\tilde{\mathbf{s}}_{k}^{a} = \mathbbm{1}(\mathbf{s}_{k}^{a} > \bar{\mathbf{s}}_{k}^{a})$, where $\bar{\mathbf{s}}_{k}^{a}$ represent the mean value of the saliency mask. Then, for the $k$-th category, we extract its region-level mean feature $\mathbf{f}(k)\in\mathbb{R}^{D_{q}}$ by MAP defined as follows:
\begin{equation}
    \mathbf{f}(k) = \frac{\sum_{\psi \in \Psi} \tilde{\mathbf{s}}_{k}^{a}\mathbf{u}_{a}(\psi) }{\sum_{\psi} \tilde{\mathbf{s}}_{k}^{a}},
\end{equation}
where $\Psi$ is the lattice of size $T \times F$. We combine the $\mathbf{f}(k)$ with the ground-truth class label
$k$
to form the anchor set 
$\mathcal{E}_{\text{anch}}=\{(\mathbf{f}(k),k)|\mathbf{f}(k)\in\mathcal{U}, k\in\mathcal{K}\}$, with $\mathcal{U}$ denoting the audio and visual feature spaces defined in Sec.~\ref{sec:method_prelim}. Since the pixel-level label $\mathbf{y}$ is available, we can directly form the visual contrastive set $\mathcal{E}_{\text{cont}}=\{(\mathbf{u}_{v}(\omega),\mathbf{y}_{v}(\omega))|\mathbf{u}_{v}(\omega)\in\mathcal{U},\mathbf{y}(\omega)\in\mathcal{Y}\}$. Hence, we can derive the positive and negative sets as:
\begin{equation}
\begin{split}
    \mathcal{P}(\mathbf{u}_{v}(\omega)) &=\{\mathbf{u}_{v}(\omega)|\mathbf{u}_{v}(\omega)\in\mathcal{U}, \mathbf{y}(\omega,k)=1\}, \\
    \mathcal{N}(\mathbf{u}_{v}(\omega)) &=\{\mathbf{u}_{v}(\omega)|\mathbf{u}_{v}(\omega)\in\mathcal{U}, \mathbf{y}(\omega,k)=0\}.
\end{split}
\end{equation}
Adopting the supervised InfoNCE~\cite{khosla2020supervised} as the objective function to pull the anchor $\mathbf{f} \in \mathcal{E}_{\text{anch}}$ 
and respective positive visual features closer while repelling anchors and their negative visual features, we define the following loss:
\begin{equation}
\scalebox{0.85}{$
\begin{aligned}
    \ell_{\text{PCL}}&(\mathbf{f}) = \frac{1}{|\mathcal{P}(\mathbf{u}_{v}(\omega))|}\sum_{\mathbf{f}_p \in\mathcal{P}(\mathbf{u}_{v}(\omega))} -\log\frac{\exp{(\mathbf{f}\cdot \mathbf{f}_p/\tau)}}{\exp{(\mathbf{f}\cdot \mathbf{f}_p/\tau)}+\sum_{\mathbf{f}_n\in\mathcal{N}(\mathbf{u}_{v}(\omega))}\exp{(\mathbf{f}\cdot \mathbf{f}_n/\tau)}},
\end{aligned}
$}
\label{eq:contrastive_loss_cavp}
\end{equation}
where $\mathbf{f}$ is an anchor feature, and $\tau$ is the temperature hyperparameter 
The combination of all sub-objective $\ell_{\text{ACP}}$, $\ell_{\text{VCP}}$ and $\ell_{\text{PCL}}$ form the CPM loss $\ell_{\text{CPM}} = \ell_{\text{ACP}} + \ell_{\text{VCP}} + \ell_{\text{PCL}}$.

\vspace{-5pt}
\subsubsection{Overall training}
The \textbf{overall training objective} is $\ell = \ell_{agn} + \lambda\ell_{\text{CPM}}$, where $\lambda$ is the weight coefficient.

\vspace{-10pt}
\section{Experiments}
\label{sec:exp}

\vspace{-5pt}
\subsection{Implementation Details}
\subsubsection{Evaluation Protocols}
We utilize standard evaluation protocols from AVSBench datasets~\cite{zhou2022audio, zhou2023audio} for single-source (SS) and multi-source (MS) scenarios with binary labels, as well as for AVSBench-Semantics with multi-class labels. Image sizes are standardized to 224 $\times$ 224 for fair comparison. 
Additionally, we adopt the original image resolution for testing, following~\cite{chen2023closer}, to demonstrate optimal model performance.
Evaluation is extended to the Visual Post-production (VPO) benchmark~\cite{chen2023closer} for challenging cases. We employ mean Intersection over Union (mIoU)\cite{everingham2015pascal} and $F_{\beta}$ score with $\beta^2=0.3$\cite{martin2004learning, zhou2022audio} to assess segmentation quality, precision, and recall performance at pixel level. Models are stratified into CNN-based pre-pixel classification and transformer-based mask classification to demonstrate architectural capabilities. 
Official training splits are used for AVSBench datasets~\cite{zhou2022audio,zhou2023audio} and VPO~\cite{chen2023closer}, with results reported on the respective testing sets. Training is performed on the entire AVSBench-Semantics dataset, and testing is conducted on subsets as well as the entire testing set to demonstrate partitioned model performance. For further details on training and inference, please refer to the \textit{Supplementary Material}.

\vspace{-10pt}
\subsubsection{Results}
We collected the experimental results from the existing benchmark~\cite{chen2023closer} and updated it with recent works~\cite{chen2024bootstrapping,li2023towards,yang2023cooperation}. 
We modified the AVSegFormer~\cite{gao2023avsegformer} with Mask2former~\cite{cheng2022masked} and denoted it as \textit{AVSegFormer*} in the tables to encourage a fair comparison with our method. 
Please note that Tab.~\ref{tab:cpm_avsbench_224_r50} includes two evaluation protocols. We employ standard semantic segmentation protocols to compute both mIoU and $F_{\beta}$ same as PascalVOC~\cite{everingham2015pascal}, initially mentioned in~\cite{zhou2022audio}.

\begin{table*}[t!]
    \centering
    \caption{Quantitative ($\mathcal{J}$, $\mathcal{F}$) audio-visual segmentation results (\%) for the AVSBench test sets~\cite{zhou2022audio,zhou2023audio} (resized to 224$\times$224) with ResNet50~\cite{he2016deep} backbone. Best results in \textbf{bold}, $2^{nd}$ best \underline{underlined}. Improvements against the $2^{nd}$ best are in the last row.
    }
    \vspace{-10pt}
    \label{tab:cpm_avsbench_224_r50}
    \def\arraystretch{1.1}
    \resizebox{1.0\linewidth}{!}{
    \begin{tabular}{!{\vrule width 1.2pt}c!{\vrule width 1.2pt}c!{\vrule width 1.2pt}c!{\vrule width 1.2pt}ccc|ccc|ccc!{\vrule width 1.2pt}} 
\specialrule{1.2pt}{0pt}{0pt}
\multirow{2}{*}{D-ResNet50~\cite{he2016deep}} & \multirow{2}{*}{Method} & \multirow{2}{*}{\makecell{External\\Model}}  & \multicolumn{3}{c|}{AVSBench-Semantics (SS)} & \multicolumn{3}{c|}{AVSBench-Semantics (MS)} & \multicolumn{3}{c!{\vrule width 1.5pt}}{AVSBench-Semantics} \\
\cline{4-12}
\rule{0pt}{10pt}
~ & ~ & ~ & 
$\mathcal{J}$\&$\mathcal{F}$ $\uparrow$ & $\mathcal{J}$ $\uparrow$ & $\mathcal{F}$ $\uparrow$ & 
$\mathcal{J}$\&$\mathcal{F}$ $\uparrow$ & $\mathcal{J}$ $\uparrow$ & $\mathcal{F}$ $\uparrow$ &  
$\mathcal{J}$\&$\mathcal{F}$ $\uparrow$ & $\mathcal{J}$ $\uparrow$ & $\mathcal{F}$ $\uparrow$ \\ 
\specialrule{1.2pt}{0pt}{0pt}
\multirow{4}{*}{\makecell{Per-pixel\\Classification}} 
& TPAVI~\cite{zhou2022audio} & \textcolor{red}{\ding{55}} & 78.80 & 72.79 & 84.80 & 52.84 & 47.88 & 57.80 & 22.69 & 20.18 & 25.20 \\ 
& AVSBG~\cite{hao2023improving} & \textcolor{red}{\ding{55}}& 79.77 & 74.13 & 85.40 & 50.88 & 44.95 & 56.80 & - & - & - \\ 
& ECMVAE~\cite{mao2023multimodal} & \textcolor{red}{\ding{55}}& 81.42 & 76.33 & 86.50 & 54.70 & 48.69 & 60.70 & - & - & - \\ 
& DiffusionAVS~\cite{mao2023contrastive} & \textcolor{red}{\ding{55}}& 81.35 & 75.80 & 86.90 & 55.94 & 49.77 & 62.10 & - & - & - \\ 
& CAVP & \textcolor{red}{\ding{55}}& 83.84 & 78.78 & 88.89 & 61.48 & 55.82 & \underline{67.14} & 32.83 & 30.37 & 35.29 \\ 
\hline
\multirow{6}{*}{Transformer} 
& CATR~\cite{li2023catr} & \textcolor{red}{\ding{55}}& 80.70 & 74.80 & 86.60 & 59.05 & 52.80 & 65.30 & - & - & - \\ 
& AuTR~\cite{liu2023audio} & \textcolor{red}{\ding{55}}& 80.10 & 75.00 & 85.20 & 55.30 & 49.40 & 61.20 & - & - & - \\ 
& AQFormer~\cite{huang2023discovering} & \textcolor{red}{\ding{55}}& 81.70 & 77.00 & 86.40 & 61.30 & 55.70 & 66.90 & - & - & - \\ 
& AVSegFormer~\cite{gao2023avsegformer} & \textcolor{red}{\ding{55}}& 80.67 & 76.54 & 84.80 & 56.17 & 49.53 & 62.80 & 27.12 & 24.93 & 29.30 \\ 
& AVSC~\cite{liu2023audiovisual} & \textcolor{red}{\ding{55}}& 81.13 & 77.02 & 85.24 & 55.55 & 49.58 & 61.51 & - & - & - \\ 
& BAVS~\cite{liu2023bavs} & \textcolor{green}{\ding{51}} & 81.63 & 77.96 & 85.29 & 56.30 & 50.23 & 62.37 & 27.16 & 24.68 & 29.63 \\ 
& AVSAC~\cite{chen2024bootstrapping} & \textcolor{red}{\ding{55}} & 81.93 & 76.90 & 86.95 & 59.88 & 53.95 & 65.81 & 27.57 & 25.43 & 29.71 \\ 
& QSD~\cite{li2023towards} & \textcolor{red}{\ding{55}} & 81.80 & 77.60 & 86.00 & \underline{61.55} & \underline{59.60} & 63.50 & - & - & - \\ 
& COMBO~\cite{yang2023cooperation} & \textcolor{green}{\ding{51}} & \underline{85.90} & \textbf{81.70} & \underline{90.10} & 60.55 & 54.50 & 66.60 & \underline{35.30} & \underline{33.30} & \underline{37.30} \\
\rowcolor{LightCyan}\cellcolor{white}
& \textbf{CPM} & \textcolor{red}{\ding{55}}  & 
\textbf{85.92} & \underline{\textbf{81.37}} & \textbf{90.47} & 
\textbf{65.40} & \textbf{59.80} & \textbf{71.00} & 
\textbf{37.05} & \textbf{34.53} & \textbf{39.57} \\ \hline
%
\specialrule{1.2pt}{0pt}{0pt}
    \end{tabular}
}
\vspace{-10pt}
    
\end{table*}

\vspace{-10pt}
\subsection{Performance on Low-resolution AVSBench Videos} 
We adopt established methodologies~\cite{liu2023audio,mao2023multimodal,liu2023bavs,chen2023closer} for conducting performance evaluations on down-sampled image benchmarks, such as AVSBench-Objects~\cite{zhou2022audio}, which includes single-source (SS) and multi-source (MS) splits with binary annotations, as well as the AVSBench-Semantics dataset~\cite{zhou2023audio}, which contains multi-class annotations. We compare the performance of state-of-the-art (SOTA) methods with our CPM in Tab.~\ref{tab:cpm_avsbench_224_r50} using mIoU and $F_{\beta}$. The results demonstrate that our model surpasses the second-best CAVP~\cite{chen2023closer} in terms of mIoU by $1.87\%$ on AVSBench-Object (SS)~\cite{zhou2022audio}, $2.83\%$ on AVSBench-Object (MS)~\cite{zhou2022audio} and $1.79\%$ on AVSBench-Semantics~\cite{zhou2022audio} using the ResNet-50~\cite{he2016deep} backbone.

\begin{table*}[t]
    \centering
    \caption{Quantitative (mIoU, $F_{\beta}$) audio-visual segmentation results (in \%) for the AVSBench-Semantic (AVSS) test sets~\cite{zhou2023audio} (original resolution) with ResNet50~\cite{he2016deep} backbone. 
    Best results in \textbf{bold}, $2^{nd}$ best \underline{underlined}. Improvements against the $2^{nd}$ best are in the last row.
    }
    \vspace{-10pt}
    \label{tab:avsbench_r50}
    \rowcolors{5}{LightCyan}{LightCyan}
    \resizebox{1.0\linewidth}{!}{
    \begin{tabular}{!{\vrule width 1.2pt}c!{\vrule width 1.2pt}c!{\vrule width 1.2pt}cc|cc|cc!{\vrule width 1.2pt}} 
\specialrule{1.2pt}{0pt}{0pt}
\multirow{2}{*}{D-ResNet50~\cite{he2016deep}} & \multirow{2}{*}{Method} & \multicolumn{2}{c|}{AVSS (SS)} & \multicolumn{2}{c|}{AVSS (MS)} & \multicolumn{2}{c!{\vrule width 1.2pt}}{AVSS} \\
\cline{3-8}
\rule{0pt}{10pt}
~ & ~ &  mIoU $\uparrow$ & $F_{\beta}$ $\uparrow$ & mIoU $\uparrow$ & $F_{\beta}$ $\uparrow$ & mIoU $\uparrow$ & $F_{\beta}$ $\uparrow$ \\ 
\specialrule{1.2pt}{0pt}{0pt}
\global\let\CT@@do@color\relax 
\multirow{2}{*}{\makecell{Per-pixel\\Classification}} 
& TPAVI~\cite{zhou2022audio} 
& 42.10 & 61.46 & 26.33 & 40.99 & 43.39 & 59.24 \\
& CAVP~\cite{chen2023closer} 
& \underline{56.91} &\underline{69.15} & \underline{38.61} & \underline{52.92} & \underline{50.75} & \underline{64.57} \\ \hline 
\multirow{2}{*}{Transformer} 
& AVSegFormer~\cite{gao2023avsegformer} 
& 46.25 & 59.76 & 27.21 & 41.38 & 41.48 & 56.21 \\ 
& AVSegFormer*~\cite{gao2023avsegformer} 
& 50.52 & 63.75 & 31.40 & 42.81 & 45.80 & 59.16 \\ 
& \global\let\CT@@do@color\oriCT@@do@color
\textbf{CPM} 
& \textbf{61.71} & \textbf{72.94} & \textbf{43.11} & \textbf{56.28} & \textbf{57.25} & \textbf{70.54} \\ \hline
Improvement & \textbf{CPM} 
& \improve{4.80} & \improve{3.79} 
& \improve{4.50} & \improve{3.36} 
& \improve{6.50} & \improve{5.97} \\
\specialrule{1.2pt}{0pt}{0pt}
    \end{tabular}
}
\vspace{-10pt}
    
\end{table*}

\begin{table*}[t]
    \centering
    \caption{Quantitative (mIoU, $F_{\beta}$) audio-visual segmentation results (in \%) for the VPO test sets (original resolution) with ResNet50~\cite{he2016deep} backbone. Best results in \textbf{bold}, $2^{nd}$ best \underline{underlined}. Improvements against the $2^{nd}$ best are in the last row.}
    \label{tab:vpo_resnet50}
    \vspace{-10pt}
    \rowcolors{5}{LightCyan}{LightCyan}
    \resizebox{1.0\linewidth}{!}{
    \begin{tabular}{!{\vrule width 1.2pt}c!{\vrule width 1.2pt}c!{\vrule width 1.2pt}cc|cc|cc!{\vrule width 1.2pt}} 
\specialrule{1.2pt}{0pt}{0pt}
\multirow{2}{*}{D-ResNet50~\cite{he2016deep}} & \multirow{2}{*}{Method} & \multicolumn{2}{c|}{VPO (SS)} & \multicolumn{2}{c|}{VPO (MS)} & \multicolumn{2}{c!{\vrule width 1.2pt}}{VPO (MSMI)} \\
\cline{3-8}
\rule{0pt}{10pt}
~ & ~ &  mIoU $\uparrow$ & $F_{\beta}$ $\uparrow$ & mIoU $\uparrow$ & $F_{\beta}$ $\uparrow$ & mIoU $\uparrow$ & $F_{\beta}$ $\uparrow$ \\ 
\specialrule{1.2pt}{0pt}{0pt}
\global\let\CT@@do@color\relax 
\multirow{2}{*}{\makecell{Per-pixel\\Classification}} 
& TPAVI~\cite{zhou2022audio} 
& 52.75 & 69.54 & 54.30 & 71.95 & 51.73 & 68.85 \\ 
& CAVP 
& \underline{62.31} & \underline{78.46} & \underline{64.31} & \underline{78.92} & \underline{60.36} & \textbf{75.60} \\ \hline
\multirow{2}{*}{Transformer} 
& AVSegFormer~\cite{gao2023avsegformer} 
& 57.55 & 73.03 & 58.33 & 74.28 & 54.22 & 70.39 \\
& AVSegFormer*~\cite{gao2023avsegformer} 
& 60.51 & 74.81 & 62.91 & 77.33 & 56.24 & 72.67 \\ 
& \global\let\CT@@do@color\oriCT@@do@color\textbf{CPM} & 
\textbf{67.09} & \textbf{79.88} & \textbf{65.91} & \textbf{79.90} & \textbf{60.55} & \underline{75.58} \\ \hline
Improvements & \textbf{CPM} 
& \improve{4.78} & \improve{1.42} 
& \improve{1.60} & \improve{0.98} 
& \improve{0.19} & \drop{-0.02} \\
\specialrule{1.2pt}{0pt}{0pt}
    \end{tabular}
}
\vspace{-10pt}
\end{table*}
\begin{table}[t]
\centering
\caption{Ablation study of the model components on AVSBench-Semantics~\cite{liu2023audio}.}
\vspace{-10pt}
\resizebox{1.0\linewidth}{!}{%
\begin{tabular}{
!{\vrule width 1.2pt}P{50pt}|P{50pt}|P{50pt}|P{50pt}|P{50pt}!{\vrule width 1.2pt}P{50pt}|P{50pt}!{\vrule width 1.2pt}
} 
\specialrule{1.2pt}{0pt}{0pt}
\multicolumn{5}{!{\vrule width 1.2pt}c!{\vrule width 1.2pt}}{Method} & \multicolumn{2}{c!{\vrule width 1.2pt}}{Metrics}  \\ \hline
Baseline & CCDM & VCP & ACP & PCL & mIoU $\uparrow$ & $F_{\beta}$ $\uparrow$ \\ \hline
\redtick & & & & & 53.04 & 65.29 \\
\redtick & \redtick & & & & 54.12 & 66.65 \\
\redtick & \redtick & \redtick & & & 55.79	& 68.17 \\
\redtick & \redtick &  & \redtick & & 55.07	& 68.03 \\
\redtick & \redtick & \redtick & \redtick & & 56.41	& 69.07 \\
\redtick & \redtick & \redtick & \redtick & \redtick & \textbf{57.25} & \textbf{70.54} \\

\specialrule{1.2pt}{0pt}{0pt}
\end{tabular}
\label{tab:ablation_components}
}
\vspace{-10pt}
\end{table}

\subsection{Performance on Original AVSBench Videos}
\begin{figure}[t!]
    \centering
    \vspace{-10pt}
    \includegraphics[width=1.0\linewidth]{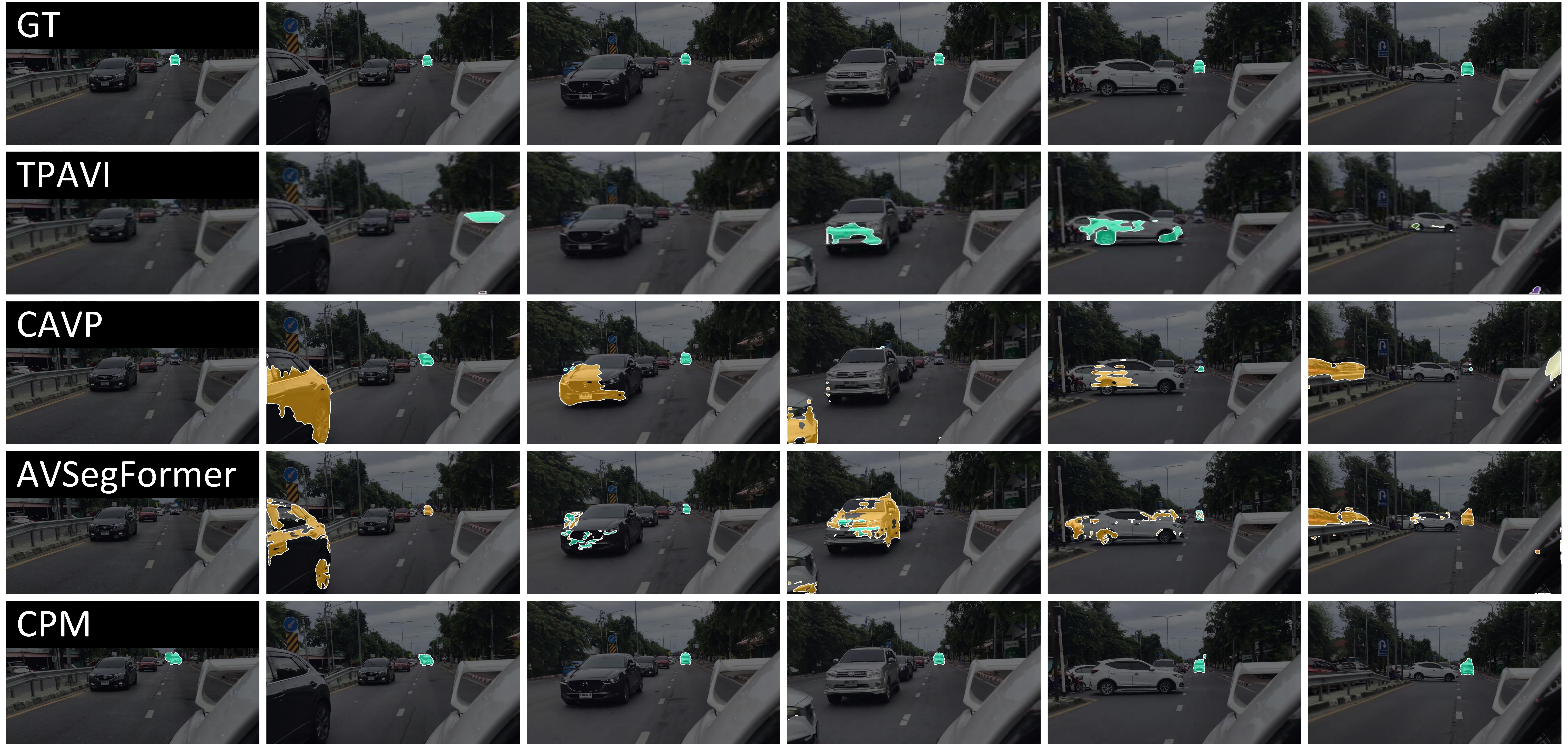}
    \vspace{-10pt}
    \caption{Qualitative audio-visual segmentation results on AVSBench-Semantics~\cite{zhou2023audio} by TPAVI~\cite{zhou2022audio}, AVSegFormer~\cite{gao2023avsegformer}, CAVP~\cite{chen2023closer} and our CPM, which can be compared with the ground truth (GT) \textcolor[HTML]{57E5C0}{Ambulance} of the first row.
    }
    \vspace{-10pt}
    \label{fig:visual}
\end{figure}

\vspace{-5pt}
The benchmark mentioned above adopted low-resolution benchmarks using significantly resized input images (from 720p to $224\times224$). While this simplifies model training, disregarding the original image aspect ratio can lead to a degradation in model performance, which is not advisable for segmentation tasks. Therefore, we follow the evaluation method outlined in~\cite{chen2023closer} for training and testing on the raw AVSBench videos. This includes using random resized crops for training and performing frame-by-frame evaluation during testing. The initial published version of AVSBench~\cite{zhou2022audio} only offers single-source and multi-source partitioning. However, this partitioning was not conducted in the later version~\cite{zhou2023audio}, resulting in some cases being overlooked. We re-organised the previous SS and MS with the newly added video data, resulting in 1278 single-source videos (AVSS-SS) and 276 multi-source videos (AVSS-MS). We compare the results with mIoU and $F_{\beta}$ to demonstrate the comprehensive model performance in Tab.~\ref{tab:avsbench_r50}. Our method shows a significant mIoU improvement of $4.80\%$ on AVSS-SS, $4.50\%$ on AVSS-MS and $+6.50\%$ on the entire AVSS dataset.
To further demonstrate the effectiveness of our method, we show a visualisation of 6-second video clip in Fig.~\ref{fig:visual} that displays a qualitative comparison between TPAVI, AVSegFormer, CAVP and our CPM. Our method can successfully approximate the ground truth segmentation of the target sound source within a group of other semantic objects.

\subsection{Performance on VPO Dataset}
We also compare the model performance on the VPO benchmark~\cite{chen2023closer}, equipped with synthesized stereo audios. We adopted ResNet-50~\cite{he2016deep} backbone for all three subsets, and the results are shown in Tab.~\ref{tab:vpo_resnet50}. To facilitate stereo audio encoding, we adopt the approach outlined in~\cite{chen2023closer} by adjusting the number of input channels in the first layer to 2.
Our approach surpasses the SOTA method CAVP~\cite{chen2023closer} by $4.78\%$ and $1.60\%$ in terms of mIoU on VPO (SS) and VPO (MS), respectively. However, we observe a marginal improvement of $0.19\%$ on the VPO (MSMI) setup compared with the SS and MS subsets. The possible explanation for this phenomenon may stem from the overly simplistic integration of stereo audio within the transformer architecture, as the foundational AVS transformer architectures~\cite{liu2023bavs,gao2023avsegformer,li2023catr,li2023towards} did not account for a distinct positional encoding scheme tailored for stereo data. We will explore the design of a dedicated stereo AVS transformer architecture as part of our future work.

\begin{figure*}[t!]
    \centering
    \begin{subfigure}[t]{0.49\textwidth}
        \centering
        \includegraphics[width=\linewidth]{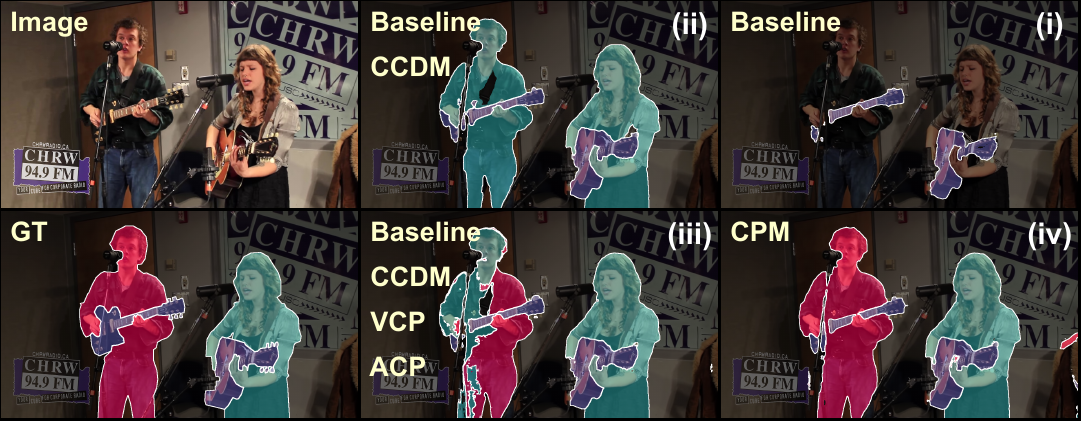}
        \caption{Visual comparison.}
        \label{fig:ablation-audio-dec-visual}
    \end{subfigure}
    \begin{subfigure}[t]{0.49\textwidth}
        \centering
        \includegraphics[width=\linewidth]{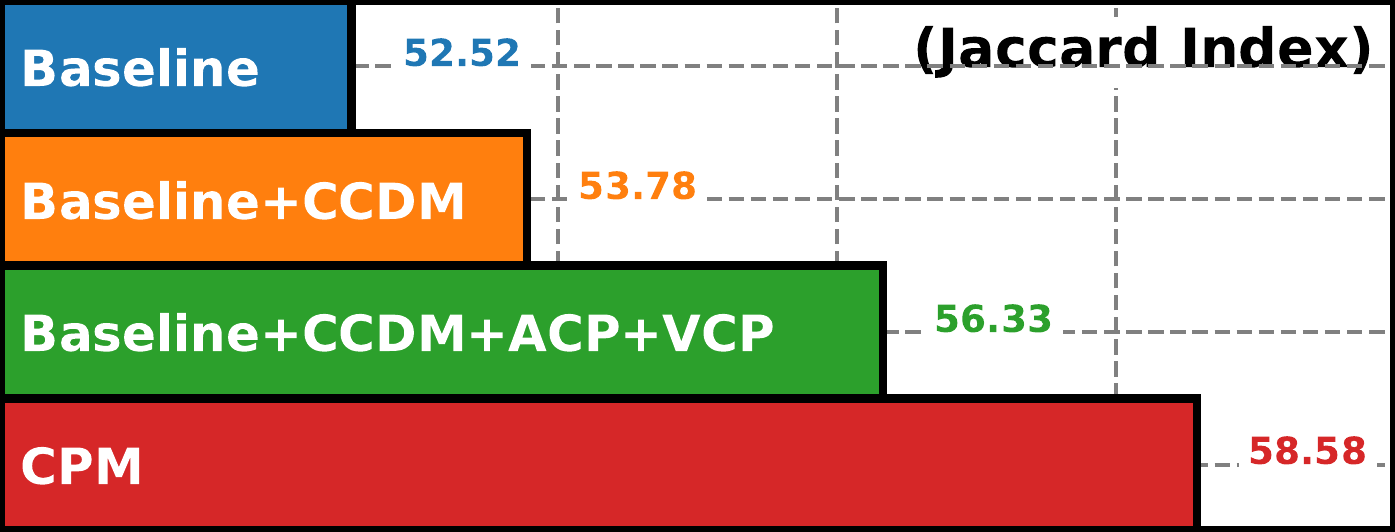}
        \caption{Coverage score comparison.}
        \label{fig:ablation-audio-dec-numerical}
    \end{subfigure}
    \vspace{-10pt}
    \caption{Qualitative (\ref{fig:ablation-audio-dec-visual}) and quantitative (\ref{fig:ablation-audio-dec-numerical}) comparisons between model components in a multi-source scenario (i.e., male singing, female singing and guitar) 
    }
    \vspace{-10pt}
    \label{fig:ablation-audio-dec}
\end{figure*}

\vspace{-10pt}
\subsection{Ablation Study}
\vspace{-5pt}
\subsubsection{Ablation of Key Components} We first perform the key components analysis of CPM on AVSBench-Semantics~\cite{zhou2023audio} in Tab.~\ref{tab:ablation_components}. The baseline is AVSegformer*~\cite{gao2023avsegformer} in the 1st row. We replace the Softmax classifier in baseline with CCDM defined in~\eqref{eq:gmm_posterior}, and observe an mIoU improvement of $+1.08\%$. By integrating the model with VCP (3rd row), using training loss in~\eqref{eq:loss_VCP}, and ACP (4th row), using loss in~\eqref{eq:loss_ACP}, separately, we achieve an improvement of $+1.26\%$ and $+1.39\%$, respectively. Subsequently, when we apply both ACP and VCP methods (5th row) the mIoU performance improves $+2.04\%$ compared to the 2nd row. 
The final row displays the complete CPM method, incorporating the dense contrastive learning approach we introduced, which leads to an additional mIoU improvement of $+1.09\%$.

\begin{figure}[t]
\centering
\begin{minipage}[b]{0.49\linewidth}
\vspace{0pt}
    \centering
    \includegraphics[width=0.9\linewidth]{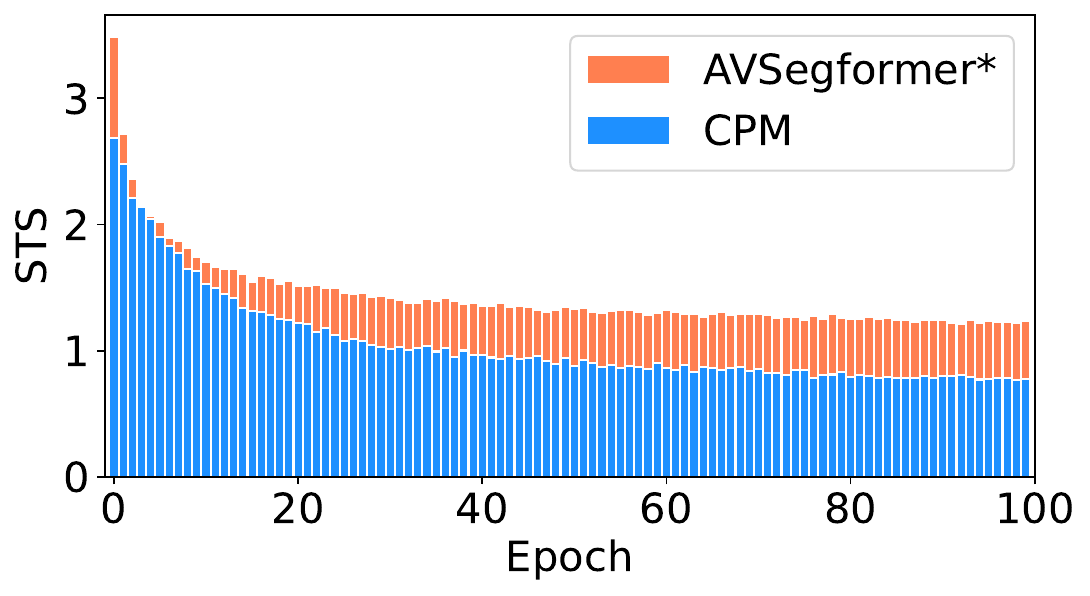}
    \vspace{-10pt}
    \captionof{figure}{Matching stability ($\mathsf{STS}\downarrow$) comparison on AVSBench-Semantics~\cite{liu2023audio}
    }
    \label{fig:stability_score}
\end{minipage}%
\hfill
\begin{minipage}[b]{0.49\linewidth}
\centering
\vspace{0pt}
\begin{table}[H]
    \caption{Ablation of GMM modelling on AVSBench-Semantics~\cite{liu2023audio}.}
    \vspace{-20pt}
    \begin{center}
    \resizebox{.8\linewidth}{!}{%
    \begin{tabular}{!{\vrule width 1.2pt}l||m{50pt}|m{50pt}!{\vrule width 1.2pt}}
        \specialrule{1.2pt}{0pt}{0pt}
        Method & mIoU $\uparrow$ & $F_{\beta}$ $\downarrow$ \\ \hline
        Point Rep. & 55.96 & 68.58 \\
        Dist Rep. & 57.25 & 70.54 \\
        \specialrule{1.2pt}{0pt}{0pt}
        $M$ & mIoU & $F_{\beta}$ \\ \hline
        1 & 55.83 & 68.68 \\
        3 & 57.25 & 70.54 \\
        5 & 56.31 & 69.04 \\
        7 & 56.05 & 68.59 \\
        \specialrule{1.2pt}{0pt}{0pt}
    \end{tabular}    
    }
    \label{tab:ablation_gmm}
    \end{center}
\end{table}
\end{minipage}%
\hspace*{\fill}
\vspace{-20pt}
\end{figure}

\vspace{-10pt}
\subsubsection{Ablation of Bipartite Matching Stability} 
To study the stability of the bipartite matching process, we design an entropy-based stability score $\mathsf{STS}$ to quantify such performance on AVSBench-Semantics~\cite{zhou2023audio}.
Intuitively, the $\mathsf{STS}$ measures the average assignment consistency after bipartite matching across all the classes.
During the training process, we collect the assigned label $\tilde{\mathbf{y}}_{i} \in \{0,..., C\}^{N}$ to the class-agnostic query $\tilde{\mathbf{q}}_{n}$ produced by the Hungarian algorithm for each training image, resulting in $\mathbf{R} = [\tilde{\mathbf{y}}_{0}\oplus\tilde{\mathbf{y}}_{1}\oplus,...,\oplus\tilde{\mathbf{y}}_{D}]$, where $\oplus$ represents the concatenation operator and $\mathbf{R}$ has size $D\times N$. We then use the indicator function to extract the assignment information over all semantic classes. We denote the resulting set as $\{\mathbf{s}_{0}, \mathbf{s}_{1}, ..., \mathbf{s}_c\}$, where $\mathbf{s}_c(n)=\sum_{d}^{D}\mathbbm{1}(\mathbf{R}(d,n)=c)$ for $n\in\{1,...,N\}$, which forms $\mathbf{s}_c \in [0,1]^N$ after normalising it to be a probability distribution. 
Finally, we calculate the stability score $\mathsf{STS} = \frac{1}{C} \sum_{c=1}^{C}H(\mathbf{s}_c)$, where $H(\cdot)$ computes the entropy. We compare our CPM with AVSegformer* over 100 training epochs as illustrated in Fig.~\ref{fig:stability_score}. Our findings reveal that our CPM consistently improves the stability score (showing lower average entropy) throughout the training process in comparison to AVSegformer*, thereby validating the efficacy of our approach in enhancing bipartite matching stability. Please refer to the \textit{Supplementary Material} for the Pseudo-code of $\mathsf{STS}$.

\vspace{-10pt}
\subsubsection{Ablation of Audio Contribution} While experimenting with the baseline method mentioned above, we empirically observed that the classification after the audio transformer decoder (i.e., $f^{\text{TD-A}}$ in Fig.~\ref{fig:frame_cpm}) has low accuracy, with noisy predictions. 
We also found that these noisy predictions are progressively refined by the vision transformer decoder layer based solely on visual clues. 
Such an observation illustrates that in some hard cases, we may fail to extract useful semantic information from audio, leading to the model being overly reliant on the visual content, which may cause erroneous testing predictions, as shown in Fig.~\ref{fig:ablation-audio-dec-visual}-(i), (ii), (iii).
To monitor the amount of valid information retained by the query following interaction with the audio branch, we introduce a classification coverage score. This score is computed by comparing the predictions generated after passing through the audio transformer decoder with the ground-truth class label, using the Jaccard index.
We show qualitative results in Fig.~\ref{fig:ablation-audio-dec-visual} for the designated scenario and quantitative results in Fig.~\ref{fig:ablation-audio-dec-numerical} for the AVSBench-Semantics testing set. The comparison starts with AVSegFormer as the baseline method with progressive addition of our proposed modules, similar to Tab.~\ref{tab:ablation_components}. 
Notice that each CPM sub-module contributes to refining key semantic information from the audio modality, as evidenced by the segmentation mask predictions and the improvement in Jaccard indexes. This shows the effectiveness of CPM.

\vspace{-10pt}
\subsubsection{Ablation of Distribution Representation}
Table~\ref{tab:ablation_gmm} provides a study of the GMM data distribution representation 
on AVSBench-Semantics~\cite{zhou2023audio}. 
The first row (Point Rep.) replaces the GMM module with a linear SoftMax for the classification output, introducing a learnable $C \times D_{q}$ feature map (i.e., $D_{q}$-dimensional feature embeddings for $C$ semantic classes for the audio and visual prompting task). 
The second row (Dist Rep.) shows our approach with the CCDM module, showing that the utilisation of the distribution modelling leads to considerable improvements of $1.29\%$ and $1.96\%$ on mIoU and $F_{\beta}$ respectively. 
The next rows of Tab.~\ref{tab:ablation_gmm} show a hyper-parameter analysis on the number of GMM components (from $M=1$ to $M=7$ components). 
Results reveal that 3 GMM components enable the best fitting for the data distribution, leading to the top performance with a maximum improvement of $1.42\%$ and $1.95\%$ on mIoU and $F_{\beta}$.

\vspace{-10pt}
\section{Discussion and Conclusion}
\vspace{-10pt}

\label{sec:con}
We introduced CPM, a new audio-visual training method designed for the Maskformer based framework to enhance bipartite matching stability and improve the efficacy of cross-modal attention for audio-visual segmentation. We proposed a class-conditional prompting learning strategy that combines class-agnostic queries with class-conditional queries, sampled from
our iteratively updated generative model. The generated class-conditional queries are utilised to probe both the magnitude spectrogram and the image feature map, aiming to remove off-the-screen noise and bypass bipartite matching to produce a more stable learning process. Lastly, we extend the class-conditional queries to a new prompting-based audio-visual contrastive learning to explicitly constrain the cross-modal representations. SOTA results on AVS benchmarks suggest that CPM can be a valuable resource for future AVS research.  
\\
\noindent\textbf{Limitations and future work.} We acknowledge that the current adaptation of stereo audio into the transformer-based method has limitations, as it encodes positional and semantic information jointly within the transformer block. This contrasts with the typical approach of separately encoding these two types of information in the transformer-based framework. Our future work will concentrate on optimizing this framework through the integration of spatial reasoning.

\appendix

\section{Training and Inference Details}
During training, we apply data augmentation for image inputs with colour jitter, horizontal flipping and random scaling between 0.5 and 2.0. We randomly crop images to $512\times512$ pixels.
We downsample the audio data to 16 kHz for durations of $1$ second~\cite{zhou2022audio} or $3$ seconds~\cite{mo2022closer} of the waveform on AVSBench and VPO. Subsequently, the resampled audio sequence is processed through the Short-Time Fourier Transform (STFT) using a 512 FFT length, a Hann window size of 400, and a hop length of 160. This results in $96 \times 256$ and $300 \times 256$ magnitude spectrograms on AVSBench and VPO, respectively.
We use the AdamW~\cite{loshchilov2017decoupled} optimizer with a weight decay of 0.0001 and a polynomial learning-rate decay 
$(1-\frac{\text{iter}}{\text{total\_iter}})^\text{power}$ with $\text{power}=0.9$. We set the initial learning rate to $0.0001$ with a mini-batch size of 16 and 100 epochs training length.
During inference, we use the resized/or original resolution with a mini-batch size of 1. We set temperature $\tau$ as 0.1 and $\lambda$ as 0.5.
We adopted ResNet-50~\cite{he2016deep} image backbones and a similar setting as~\cite{cheng2022masked} for the transformer decoder blocks in the segmentation head. For the audio backbones, we use VGGish~\cite{hershey2017cnn} (following~\cite{zhou2022audio}) and ResNet-18~\cite{he2016deep} (following~\cite{chen2021localizing, mo2022closer}) for AVSBench and VPO, respectively.

\begin{algorithm}[t!]
\caption{Stability Score (STS)}
\label{alg:sts}
\definecolor{codeblue}{rgb}{0.25,0.5,0.5}
\definecolor{codekw}{rgb}{0.85, 0.18, 0.50}
\definecolor{ao}{rgb}{0.0, 0.5, 0.0}
\begin{algorithmic}[1]
    \State \textcolor{ao}{\# $N$: number of queries}
    \State \textcolor{ao}{\# $C$: number of classes}
    \State \textcolor{ao}{\# $f_{\text{Hungarian}}$: Hungarian algorithm}
    \State \textcolor{ao}{\# $H(\cdot)$: the function used to compute the negative entropy score}
    \State \textbf{require:} Training set $\mathcal{D}$ with $D$ samples, model $f_{\theta}$, class-agnostic query $\mathbf{q}$, an empty list $R$.
    \State \textcolor{ao}{\# Get assigned labels for each sample and each query.}
    \For{$(\mathbf{x}_i,\mathbf{y}_i) \in \mathcal{D}$}
        \State $\mathbf{p} = f_{\theta}(\mathbf{x}_i)$
        \State $\tilde{\mathbf{y}}_i = f_{\text{Hungarian}}(\mathbf{p})$ \textcolor{ao}{\# $\tilde{\mathbf{y}}_i: 1 \times N$ }
        \State $R$.append($\tilde{\mathbf{y}}_i$)
    \EndFor
    \State \textcolor{ao}{\# Concatenate all the assigned labels}
    \State $\mathbf{R} = \mathsf{Concat}(R)$ \textcolor{ao}{\# $\mathbf{R}: D \times N$}
    \State $\mathsf{STS} = 0$
    \State \textcolor{ao}{\# Iterate through all the classes and compute the average $\mathsf{STS}$ score.}
    \For{$c$ in $C$}
        \State $\mathbf{s}_{c} = (\mathbf{R} == c)$.sum(0) \textcolor{ao}{\# $\mathbf{s}_{c}: 1 \times N$}
        \State $\mathbf{s}_{c} = \text{clamp}(\mathbf{s}_{c}/\text{sum}(\mathbf{s}_{c}), \text{min}=1e-12, \text{max}=1)$
        \State $\mathsf{STS} = \mathsf{STS} + H(\mathbf{s}_{c})$
    \EndFor
    \State $\mathsf{STS} = \mathsf{STS} / C$
\end{algorithmic}
\end{algorithm}

\begin{figure}[ht]
    \centering
    
    \begin{subfigure}[b]{.49\linewidth}
         \centering    
            \includegraphics[width=1.\linewidth]{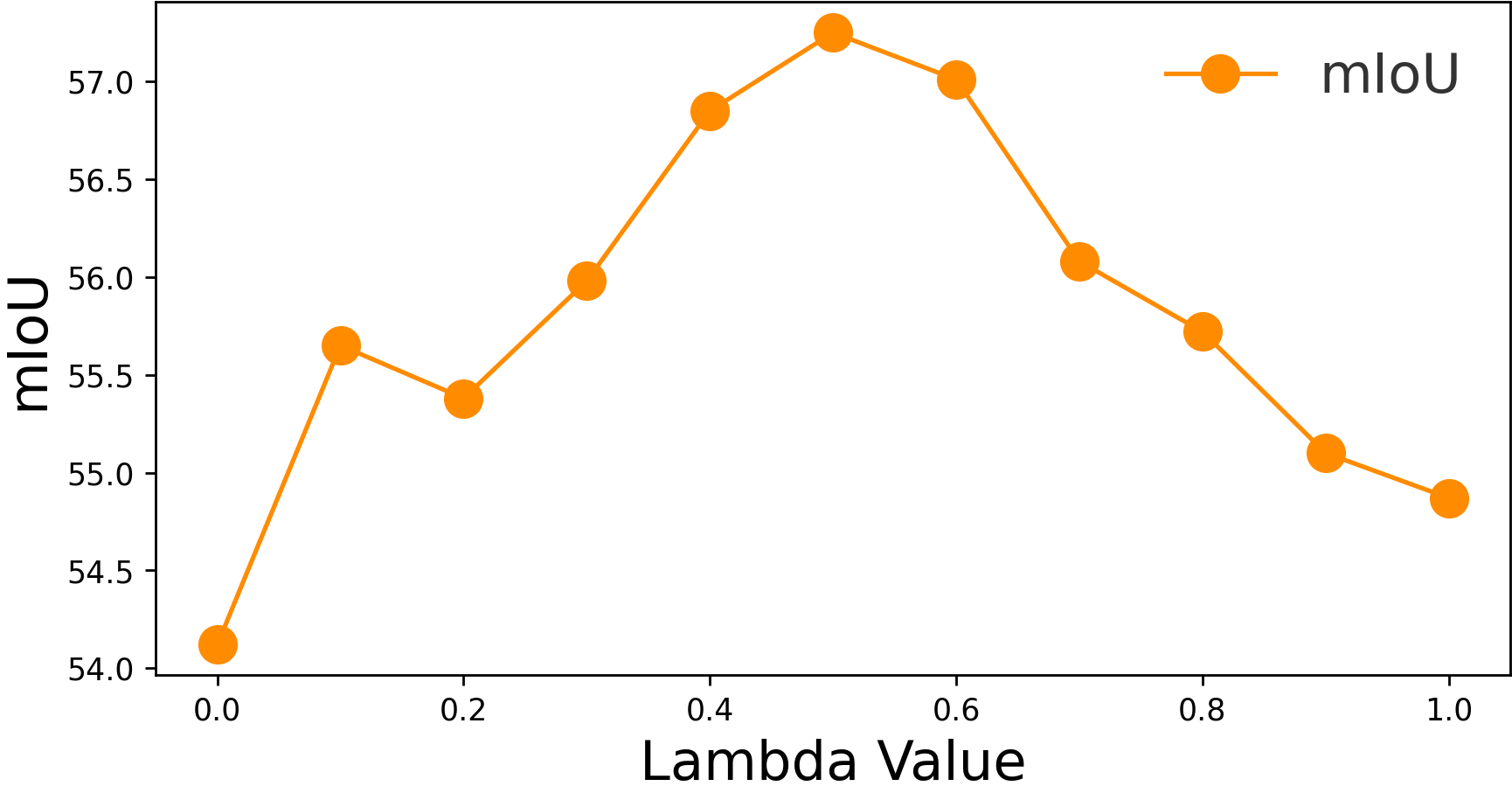}
        \caption{mIoU $\uparrow$}
        \label{fig:hyper_miou}
         
    \end{subfigure}
    \begin{subfigure}[b]{.49\linewidth}
        \centering    
            \includegraphics[width=1.\linewidth]{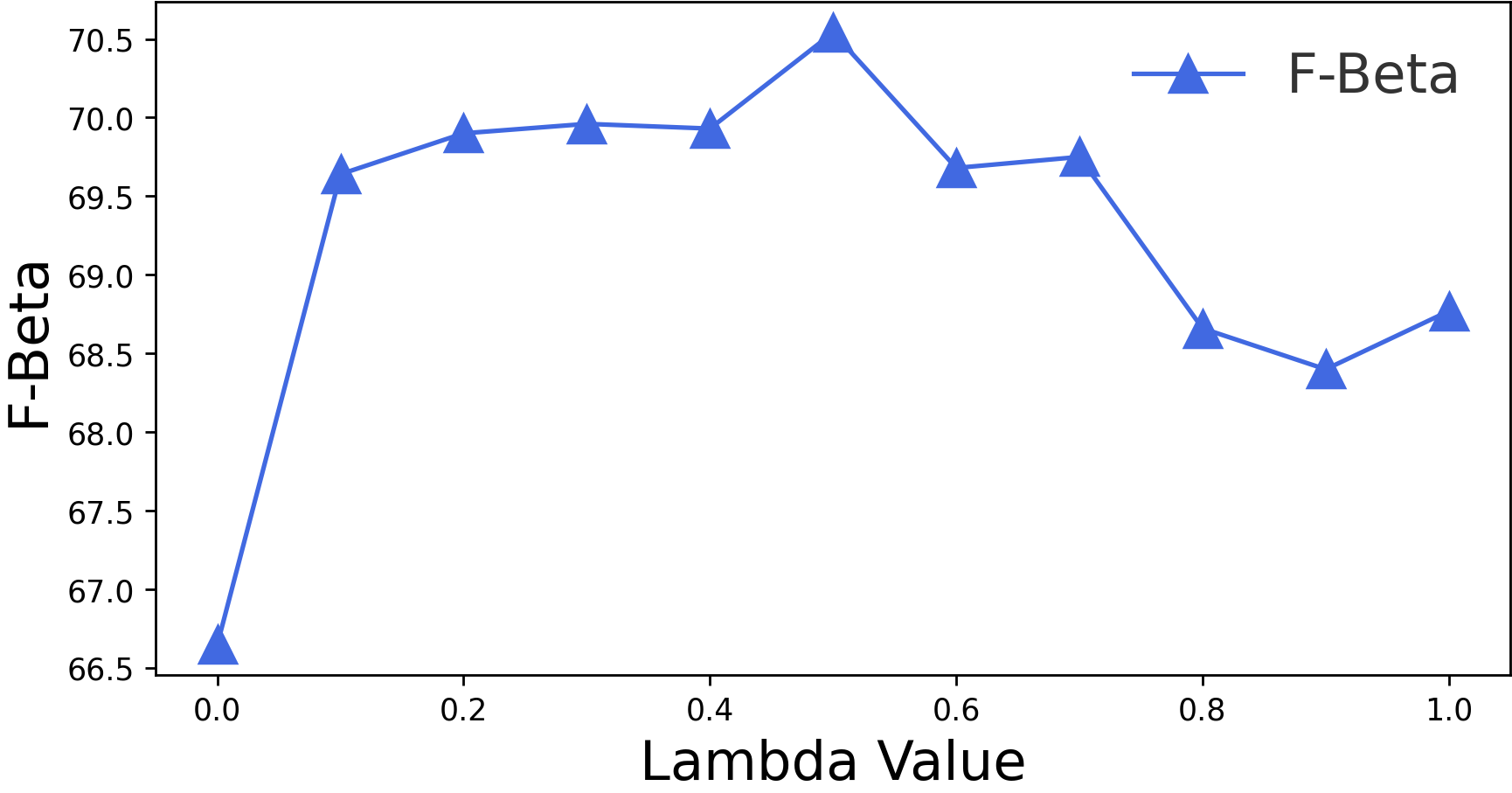}
        \caption{$F_{\beta}$ $\uparrow$}
        \label{fig:hyper_fscore}
         
    \end{subfigure}
    \caption{Ablation study on the weight coefficient $\lambda$ using the AVSBench-Semantics dataset~\cite{zhou2023audio}, with the ResNet50~\cite{he2016deep} backbone.}
    \label{fig:ablation_hyper}
    
\end{figure}

\subsection{Additional Results}

\begin{figure}[ht!]
    \centering
    \includegraphics[width=1.\linewidth]{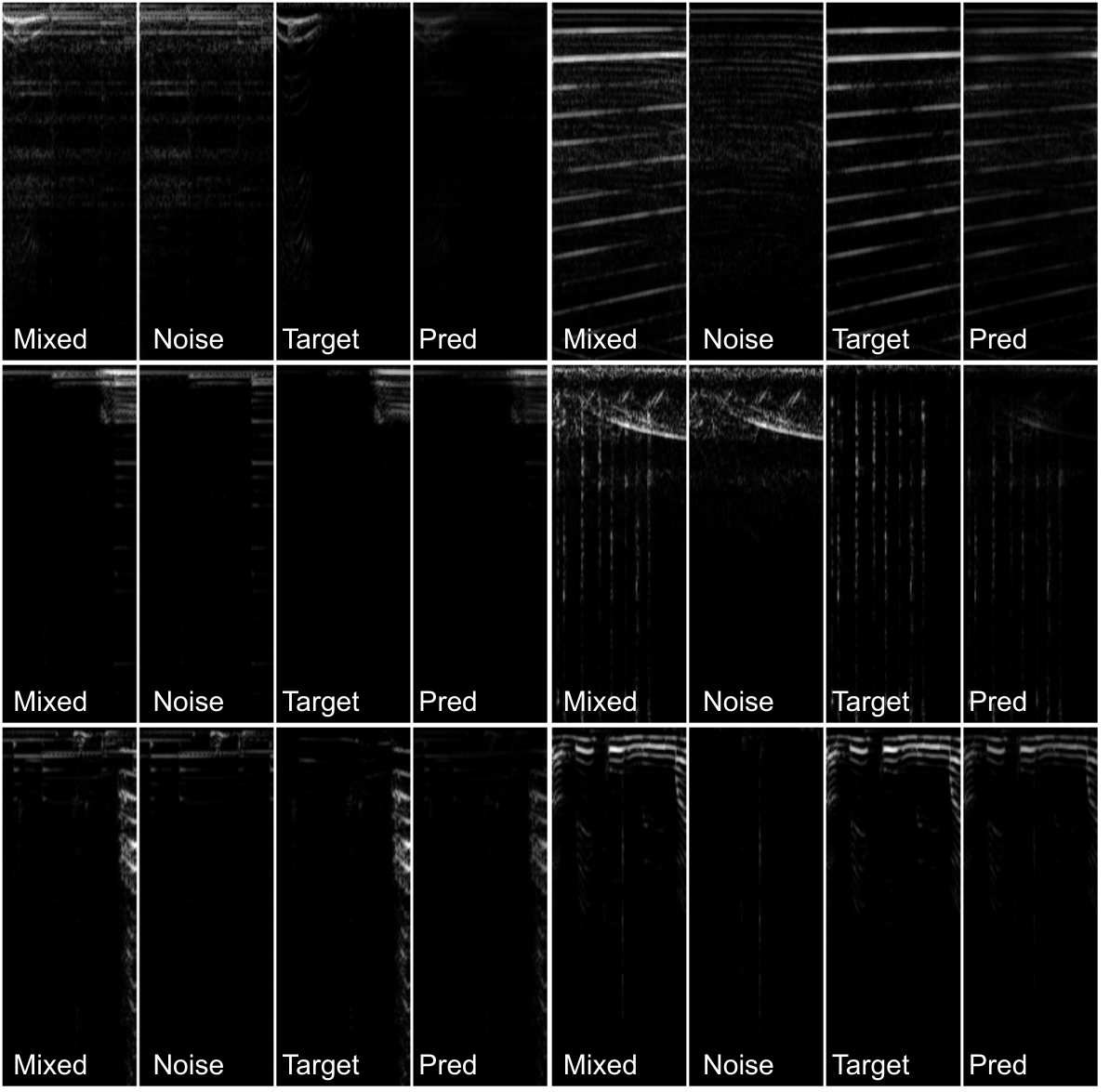}
    
    \caption{Visualization of Audio Conditional Prompting (ACP) process on six samples from the AVSBench-Semantics dataset~\cite{liu2023audio} (two examples per row). Each example includes a mixture of magnitude spectrogram (Mixed), noise beyond the visible range (Noise), the ground-truth target (Target), and the prediction generated by the CPM model (Pred).}
    \vspace{-20pt}
    \label{fig:visual_acp}
\end{figure}

\begin{figure}[htp!]
    \centering
    \includegraphics[width=0.96\linewidth]{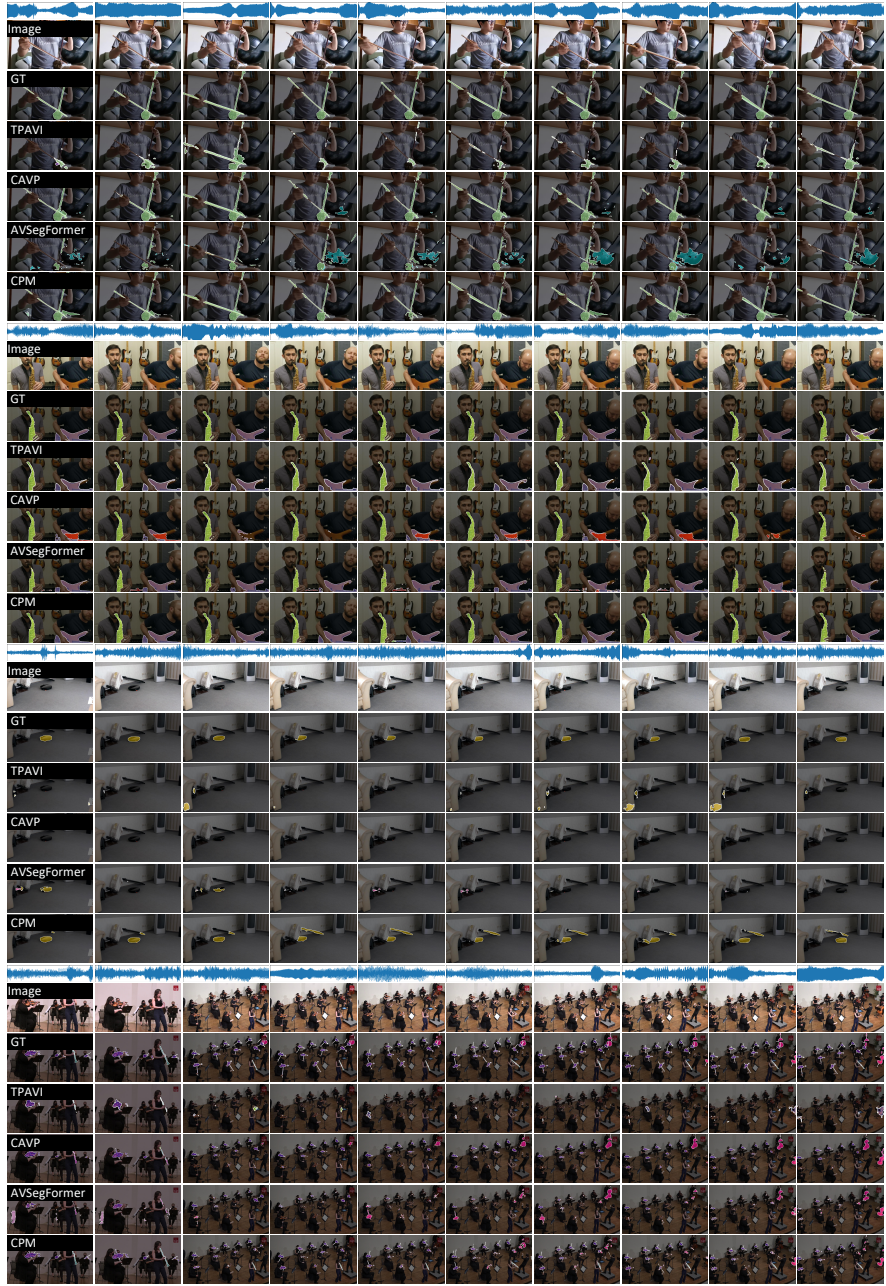}
    \caption{Qualitative audio-visual segmentation results on AVSBench-Semantics~\cite{zhou2023audio} by TPAVI~\cite{zhou2022audio}, AVSegFormer~\cite{gao2023avsegformer}, CAVP~\cite{chen2023closer} and our CPM. The prediction results can be compared with the original frame and the ground truth (GT) of the first two rows of each video.}
    \label{fig:cpm_supply_visual}
\end{figure}

\subsubsection{Average Pooling Vs. Max Pooling}
We employed masked average pooling (MAP) in our prompting-based contrastive learning (PCL) module to extract correlated audio features. While an alternative option, masked max pooling (MMP), may seem viable, we argue it is less effective due to its sensitivity to false activation and incapability to capture spatially distributed information. To validate our choice, we conducted an ablation study replacing MAP with MMP in PCL during training and compared $F_{\beta}$ scores on AVSBench-Semantics~\cite{liu2023audio}. The results showed a $1.07\%$ performance drop with MMP, indicating the effectiveness of MAP.

\begin{table}[t!]
    \centering
    \begin{tabular}{!{\vrule width 1.2pt}l!{\vrule width 1.2pt}l!{\vrule width 1.2pt}l!{\vrule width 1.2pt}l!{\vrule width 1.2pt}l!{\vrule width 1.2pt}l!{\vrule width 1.2pt}}
    \specialrule{1.5pt}{0pt}{0pt}
        ~ & \multicolumn{2}{c!{\vrule width 1.2pt}}{Per-pixel Classification} & \multicolumn{2}{c!{\vrule width 1.2pt}}{Transformer} \\ 
        \specialrule{1.5pt}{0pt}{0pt}
        Methods & TPAVI~\cite{zhou2022audio} & CAVP~\cite{chen2023closer} & AVSegFormer~\cite{gao2023avsegformer} & CPM \\ \hline
        Training & 48 h & 22 h & 23 h & 30 h \\ \hline
        Inference & 9 fps & 20 fps & 7 fps & 14 fps \\ 
    \specialrule{1.5pt}{0pt}{0pt}
    \end{tabular}
    \captionof{table}{Efficiency comparison for training and inference on AVSBench-Semantic~\cite{liu2023audio} using a ResNet50 backbone at original resolution.}
    \label{tab:infer_fps}
\end{table}

\subsection{Evaluation of Training and Inference Efficiency}
We measure the overall training time and frame per second (FPS) required for inference 
in Tab.~\ref{tab:infer_fps}. The experiments are conducted on the AVSBench-Semantic (AVSS) dataset~\cite{zhou2023audio} using a ResNet50 backbone at the original resolution. 
During inference, we evaluate models on 10 videos, sampling 100 frames per video (\ie, a total of $10\times100$ frames) to compute the number of frames per second (FPS).
The results show that our CPM requires additional training time (e.g., $+8$ hours) and has a slower inference speed (i.e., $-6$ fps) compared to CAVP. However, CPM achieves a $+7$ fps faster inference speed than AVSegFormer for a similar model architecture.

\subsubsection{Hyper-parameter Analysis}
We further conduct an ablation study to investigate the sensitivity of the hyper-parameter ($\lambda$) on AVSBench-Semantics~\cite{liu2023audio}, as depicted in Fig.~\ref{fig:ablation_hyper}. Our findings demonstrate that a moderate value of $\lambda$ is conducive to model training, whereas excessively small values ($\lambda=0.1$) may result in learning stagnation, and overly large weights ($\lambda=1.0$) could potentially be influenced by noisy gradients during the initial stages of learning.

\section{Pseudo-code for Stability Score}
We present a detailed explanation of the computation methodology for the stability score $\mathsf{STS}$, as outlined in Alg.\ref{alg:sts}. The $\mathsf{STS}$ is calculated after each epoch to evaluate the stability of the bipartite matching process through an average entropy score across all classes. The entire process comprises two main steps. Initially, we gather all assigned labels (generated by the Hungarian Algorithm\cite{carion2020end}) for each training sample. Subsequently, we iterate through all classes and compute the respective $\mathsf{STS}$ based on the negative entropy of the probability distribution of the assignments.

\section{Audio Conditional Prompting Visualisation}
We show six training examples sourced from the AVSBench Semantics~\cite{liu2023audio} dataset, as depicted in Figure \ref{fig:visual_acp}. The results demonstrate our successful utilization of class conditional prompts to search for corresponding semantic information within the perturbed magnitude spectrogram.


\bibliographystyle{splncs04}
\bibliography{egbib}
\end{document}